\documentclass[10pt,twocolumn,letterpaper]{article}

\usepackage{cvpr}              % To produce the CAMERA-READY version
\usepackage{multirow}
\usepackage{tabularx}
\definecolor{cvprblue}{rgb}{0.21,0.49,0.74}
\usepackage[pagebackref,breaklinks,colorlinks,allcolors=cvprblue]{hyperref}

\def\paperID{20198} % *** Enter the Paper ID here
\def\confName{CVPR}
\def\confYear{2026}

\title{MoReFun: Past-\underline{Mo}vement Guided Motion \underline{Re}presentation Learning for \underline{F}uture Motion Prediction and \underline{Un}derstanding}

\author{%
  Junyu Shi\textsuperscript{1}, 
  Haoting Wu\textsuperscript{1}, 
  Zhiyuan Zhang\textsuperscript{1}, 
  Lijiang Liu\textsuperscript{1}, 
  Yong Sun\textsuperscript{1}, 
  Qiang Nie\textsuperscript{1}\thanks{Corresponding author}\\
  \textsuperscript{1}The Hong Kong University of Science and Technology (Guangzhou)\\
  {\footnotesize \texttt{\{jshi890, hwu393, zzhang400, lliu135, ysun691,\}@connect.hkust-gz.edu.cn}\quad
  \texttt{\{qiangnie\}@hkust-gz.edu.cn}}
}

\begin{document}
\maketitle
\begin{abstract}
3D human motion prediction aims to generate coherent future motions from observed sequences, yet existing end-to-end regression frameworks often fail to capture complex dynamics and tend to produce temporally inconsistent or static predictions—a limitation rooted in representation shortcutting, where models rely on superficial cues rather than learning meaningful motion structure. We propose a two-stage self-supervised framework that decouples representation learning from prediction. In the pretraining stage, the model performs unified past–future self-reconstruction, reconstructing the past sequence while recovering masked joints in the future sequence under full historical guidance. A velocity-based masking strategy selects highly dynamic joints, forcing the model to focus on informative motion components and internalize the statistical dependencies between past and future states without regression interference. In the fine-tuning stage, the pretrained model predicts the entire future sequence, now treated as fully masked, and is further equipped with a lightweight future-text prediction head for joint optimization of low-level motion prediction and high-level motion understanding. Experiments on Human3.6M, 3DPW, and AMASS show that our method reduces average prediction errors by 8.8\% over state-of-the-art methods while achieving competitive future-motion understanding performance compared to LLM-based models. Code is available at: \url{https://github.com/JunyuShi02/MoReFun}
\end{abstract}    
\section{Introduction}
\label{sec:intro}

\begin{figure}[t!]
\centering
\includegraphics[width=0.43\textwidth]{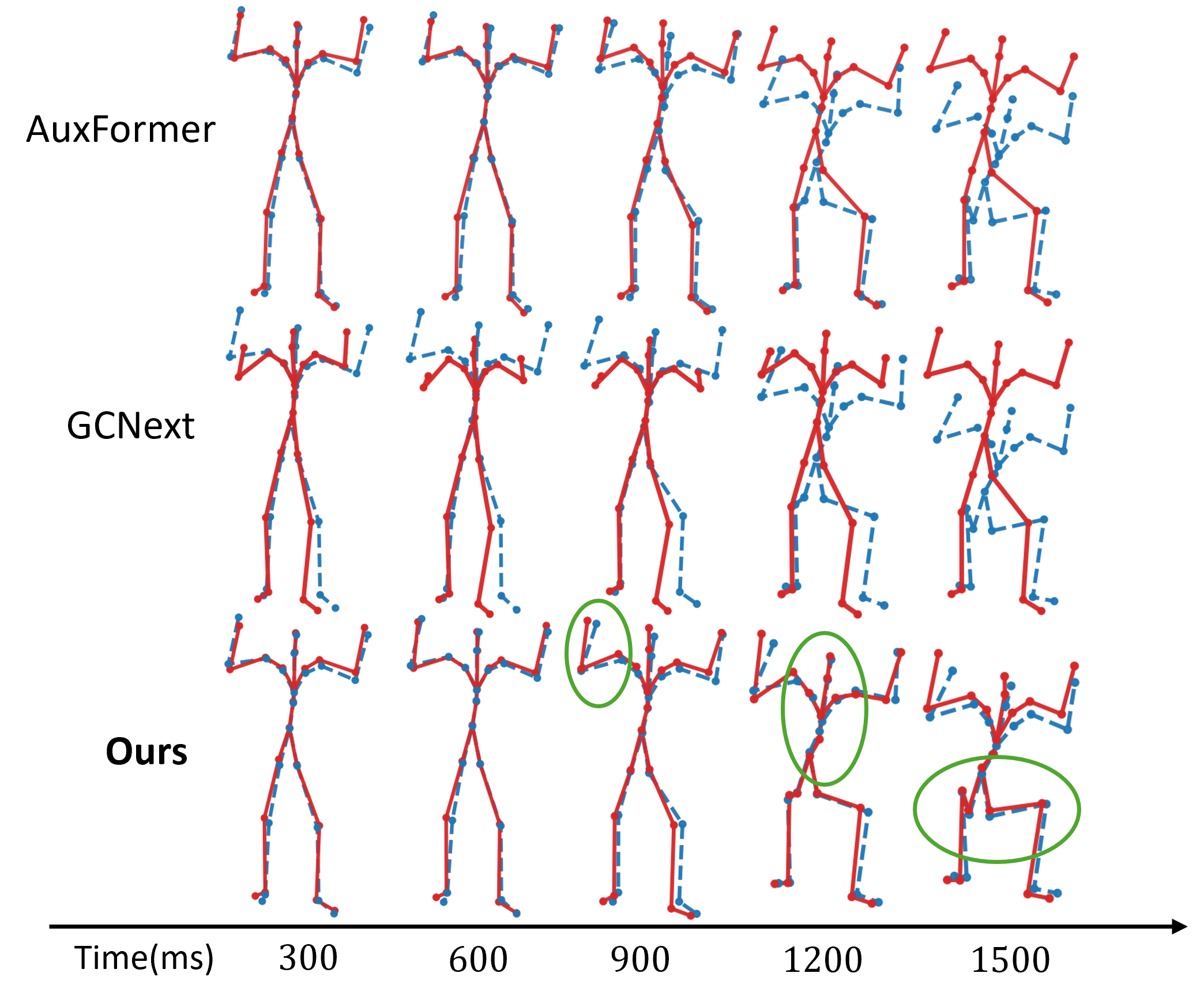}
\caption{Comparison of complex motion prediction results. While previous methods struggle with complex movements, producing poor and unnatural static poses, our method demonstrates improved accuracy by addressing the inherent weakness in representation capability, yielding a natural motion prediction.}
\label{fig:example}
\vspace{-0.5cm}
\end{figure}

3D human pose prediction, which aims to generate continuous and natural future motions based on historical observed sequences, is a pivotal technology for applications such as human-computer interaction and autonomous driving. Despite considerable strides made by recent studies, particularly those based on GNNs~\cite{ma2022progressively, cai2020learning, fu2023learning, mao2021multi, shi2023multi} and Transformers~\cite{aksan2021spatio, guo2023back, wei2025alien}, existing state-of-the-art (SOTA) models exhibit fundamental failures in accurately predicting complex motions. Their predicted motions often rapidly degenerate into unnatural static poses or lose temporal coherence, as shown in the Fig.~\ref{fig:example}.

This failure is not attributable to specific models (e.g. Transformers) but rather stems from a fundamental flaw in the prevailing end-to-end regression paradigm used for motion prediction. Current methods frame this task as a direct regression problem. This approach compels the model to simultaneously learn complex spatio-temporal representations and regress a highly uncertain future within a single training stage. This results in a phenomenon we term ``Representation Shortcutting'' where the model learns only superficial features sufficient to minimize the regression loss on the training data, while failing to capture the underlying physical principles and joint constraints governing motion.

To improve the prediction performance, researchers have pursued two main paths. The first path involves refining the regression paradigm. For instance, feed-forward architectures~\cite{ma2022progressively, xu2023eqmotion, sun2024moml} have been used to mitigate temporal error accumulation, while other methods model low-frequency DCT signals to reduce motion jitter~\cite{wang2024gcnext, li2021skeleton}. Moreover, the system latency~\cite{sun2025lal} and unexpected perturbation~\cite{yue2024human} are also under consideration. However, these methods remain constrained by the ``Representation Shortcutting'' problem. The second path explores auxiliary learning, such as using parallel denoising tasks to aid prediction~\cite{xu2023auxiliary}. Yet, this parallel multi-task paradigm has critical flaws: (1) The auxiliary and main (motion prediction) tasks may suffer from mutual interference; and (2) It fails to explicitly model the fundamental dependency of future motion on past observations, which is central to accurate motion prediction.

In this work, we decouple representation learning from the prediction task into two distinct stages. In the first stage, we propose a self-supervised learning (SSL) framework based on past-guided future reconstruction to pretrain the motion model. We mask joints within the future motion sequence and compel the model to leverage the complete past sequence as guidance to reconstruct the masked elements. This SSL approach enables the model to acquire rich motion priors, free from the interference of a regression loss, helping it grasp the transition logic and statistical dependencies between past and future states. Concurrently, we also perform self-reconstruction on the past sequence to enhance the encoder's representational power.

To maximize the efficacy of this SSL stage, we introduce a velocity-based masking strategy. Unlike random masking, we prioritize masking joints that exhibit the most significant movement between adjacent frames. This forces the model to focus its representational capacity on the most informative and dynamic components of the motion (e.g., the upper limbs during an ``eating'' action), rather than on low-information, static joints.

In the second stage, we fine-tune the pretrained model on the downstream motion prediction task. Here, we treat the entire future sequence as ``masked'' and task the model with its recovery. Furthermore, we explore the synergistic relationship between motion prediction and motion understanding. We design a novel joint fine-tuning objective where the model, given the history, is tasked with simultaneously predicting: (1) the future motion sequence (low-level prediction) and (2) a future motion description (high-level understanding). We find that these two tasks are mutually beneficial: the high-level, textual understanding of the future movement constrains and refines the low-level motion prediction, and vice versa.

Our key contributions can be summarized as follows:

\begin{itemize}

    \item We propose a self-supervised motion representation learning framework that unifies past and future self-reconstruction, with future reconstruction guided by historical motion cues. A velocity-based masking strategy is applied to emphasize dynamic joints during reconstruction. This paradigm enables the model to capture intrinsic spatiotemporal structure as well as the dependencies between past and future states.
    \item We develop a joint fine-tuning objective that tasks the model with predicting both the future motion sequence and future textual description. This coupling of low-level motion prediction and high-level semantic understanding yields mutually reinforcing improvements in coherence and realism.
    \item We conduct extensive experiments and demonstrate that our method reduces the average prediction errors by 8.8\% compared to \textit{state-of-the-art} methods on the 3DPW, Human3.6M, and AMASS datasets. Moreover, with a lightweight text prediction head, our model achieves competitive performance in future motion understanding compared to LLM-based methods~\cite{wu2025mg}.  

\end{itemize}
\section{Related Work}
\label{sec:related_work}

\begin{figure*}[t!]
\centering
\includegraphics[width=0.95\textwidth]{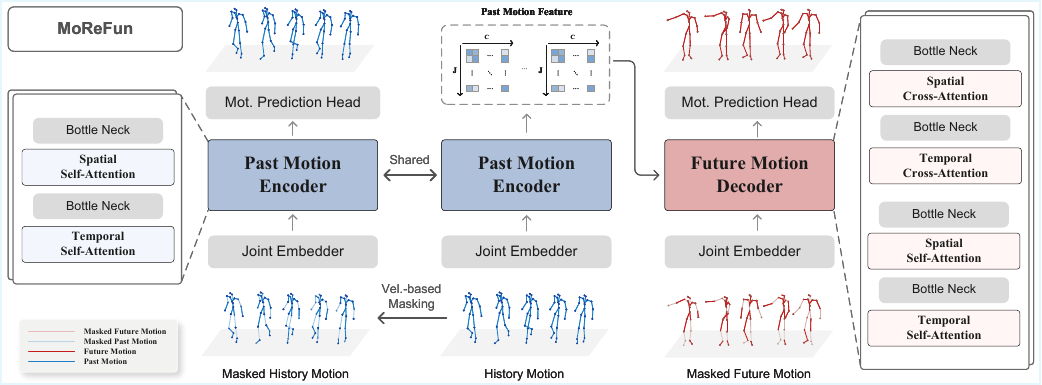}
\caption{Overall architecture of MoReFun, which consists of a Past Motion Encoder (PME) and a Future Motion Decoder (FMD). The PME is built with stacked Spatial and Temporal Self-Attention layers, processing the past motion sequence. The FMD receives masked future motion together with the encoded past-motion features, and is composed of Spatial/Temporal Cross-Attention followed by Spatial/Temporal Self-Attention. Through these stages, the decoder leverages the past-motion representation to guide the reconstruction of masked future joints.}
\label{fig:pipepine}
\vspace{-0.5cm}
\end{figure*}

\subsection{3D Skeleton-based Human Motion Prediction}

Early motion prediction methods relied on probabilistic models~\cite{wang2007gaussian, lehrmann2014efficient}, yet their strong assumptions limited their ability to capture complex human dynamics. With deep learning, RNN-based approaches~\cite{ghosh2017learning, fragkiadaki2015recurrent, habibie2017recurrent, martinez2017human} became dominant by encoding historical trajectories and autoregressively generating future poses, though they often suffer from error accumulation during long-term prediction. To overcome this, recent works shift toward feed-forward architectures that directly predict future sequences in a non-autoregressive manner. Graph-based models~\cite{ma2022progressively, li2021multiscale, dang2021msr, li2021symbiotic, shi2023multi} leverage skeletal topology, frequency-domain representations~\cite{mao2019learning, mao2020history, shi2024gradient} provide compact trajectory modeling, and dynamic GNNs~\cite{li2020dynamic, fu2023learning} adapt joint relations to motion changes. Transformer-based models~\cite{aksan2021spatio, martinez2021pose, xu2023auxiliary} further enhance long-range dependency modeling, with AuxFormer~\cite{xu2023auxiliary} improving prediction using auxiliary denoising and reconstruction tasks. Meanwhile, lightweight MLP designs~\cite{bouazizi2022motionmixer, guo2023back} demonstrate that strong performance can also be achieved without heavy attention mechanisms. Together, these feed-forward approaches alleviate autoregressive drift and have become the mainstream paradigm for modern human motion prediction.

\subsection{Self-supervised Pretraining}
Self-supervised pretraining has brought about a revolution in the understanding and generation of language and images. This technique serves as one of the fundamental training methods of Large Language Models (LLMs), which is widely used in BERT~\cite{devlin2018bert}, RoBERTa~\cite{liu2019roberta} and other models. For image processing tasks, recovering masked pixels/tokens is a common self-supervised pretraining method~\cite{he2022masked, chen2024context}. MAE~\cite{he2022masked} proves that even when a substantial portion of image information is masked, the network can reconstruct a complete image based on a limited number of visible features. The self-learning paradigm of autoencoder in MAE is also widely used in multimodal~\cite{chen2023pimae}, video processing~\cite{tong2022videomae}, motion generation~\cite{shi2025genm}, point cloud data processing~\cite{zhang2023learning} and other tasks.

\subsection{Motion Captioning}
Motion captioning aims to generate natural language descriptions for 3D human motion sequences~\cite{petrovich2022temos, shi2025mogic}. Recent LLM-based approaches~\cite{li2025lamp, jiang2023motiongpt} further leverage powerful pretrained language models to produce more accurate and semantically coherent motion descriptions. Despite these advances, the role of future-motion semantics in enhancing motion prediction remains largely underexplored, leaving a gap between motion understanding and predicting.
\section{Method}
\label{sec:method}

\subsection{Problem Background}

3D skeleton-based human motion prediction aims to generate unseen and consecutive future motions from historical observations. Given a past motion sequence $\mathbf{X}^P=[\mathbf{x}_1,\ldots,\mathbf{x}_T] \in \mathbb{R}^{T \times J \times K}$, where $J$ denotes the number of joints and $K$ is the feature dimension (typically $K=3$ for the $x,y,z$ coordinates), the goal is to predict the future motion sequence $\mathbf{X}^F=[\mathbf{x}_1,\ldots,\mathbf{x}_L] \in \mathbb{R}^{L \times J \times K}$. We therefore design a predictive model $\mathcal{P}(\cdot)$ that maps the observed past motion to its future continuation, i.e., $\mathbf{X}^F = \mathcal{P}(\mathbf{X}^P; \pi_\theta)$ with trainable parameters $\pi_\theta$. Beyond pose prediction, we further consider the generation of a semantic description of the future behavior. Given the same historical sequence, the model outputs a future text sequence $\mathbf{Y}^F=[y_1,\ldots,y_D]$ that summarizes the expected motion. This formulation naturally extends the task to a joint prediction problem, where the model produces both future motion and its corresponding semantic description: $(\mathbf{X}^F, \mathbf{Y}^F) = \mathcal{P}(\mathbf{X}^P; \pi_\theta)$.

\begin{figure}[t!]
\centering
\includegraphics[width=0.43\textwidth]{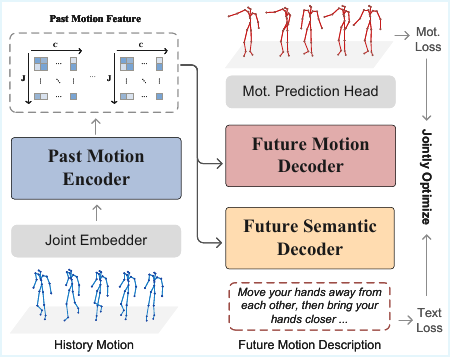}
\caption{Overview of the finetuning stage. The Past Motion Encoder extracts spatiotemporal representations from the observed history motion. The shared past-motion feature is fed into two branches: (1) the Future Motion Decoder, which predicts the future 3D pose sequence from a zero-initialized state, and (2) the Understanding Decoder, which produces a natural-language description of the upcoming motion. Both branches are jointly optimized with motion and text supervision.}
\label{fig:finetune}
\vspace{-0.5cm}
\end{figure}

\subsection{Model Architecture}
As shown in Fig.~\ref{fig:pipepine}, our proposed model consists of a Past Motion Encoder (PME) and a Future Motion Decoder (FMD). The PME aims to extract spatiotemporal features from the past motion sequence. The FMD aims to predict the future motion sequence based on the initial state (which consists of zero-valued inputs in motion prediction tasks) and past motion features.

\paragraph{Joint Embedder and Predict Head} Given an input sequence \( \mathbf{X} \), we first use a single linear layer to embed each joint of each frame. To distinguish the spatiotemporal order information of the joints, we add trainable temporal and spatial position encoding to the motion embedding. This process can be represented as:

\begin{equation}
    \mathbf{H} = \mathbf{W}\mathbf{X} + \mathbf{b} + \mathbf{P}_t + \mathbf{P}_s
\end{equation}

where \( \mathbf{W} \in \mathbb{R}^{K \times C}, \mathbf{b} \in \mathbb{R}^C \) are the weights and biases of the linear layer, and \( \mathbf{P}_t, \mathbf{P}_s \) are the temporal and spatial position encoding, which are randomly initialized and trainable. \( C \) denotes the number of feature channels. For the output of PME and FMD, a linear layer as the Predict Head is applied to obtain the 3D motion sequence.

\paragraph{Past Motion Encoder (PME)} We design PME to model the features of the past motion sequence. Human motion sequences contain two features: the change of joint position over time and the relationship between joints in space. We use the spatiotemporal attention to simultaneously model the spatiotemporal information of the joints. Specifically, PME contains \( N \) cascaded spatiotemporal attention modules, where each module first models the sequence dependencies in the temporal dimension and then models the joint relationships in the spatial dimension. Assuming the input to the \( l^{th} \) layer is \( \mathbf{H}^l \), the computation process can be expressed as:

\begin{align}
    \mathbf{H}^{l+1} = \mathcal{F}_{attn}^s(\mathcal{F}_{attn}^t(\mathbf{H}^l))
\end{align}

where \( \mathcal{F}_{attn}^s(\cdot) \) / \(\mathcal{F}_{attn}^t(\cdot) \) represents the spatial/temporal multi-head self-attention operation with the bottle neck.

\paragraph{Future Motion Decoder (FMD)} The purpose of FMD is to generate future motion sequences based on the initial state and the output from PME (i.e., past motion features). FMD comprises two modules: the spatiotemporal attention module for modeling the sequence dependencies within itself and the spatiotemporal cross-attention to incorporate past motion features into the current state, which can be mathematically represented as:

\begin{align}
    \mathbf{H}^{l+1} = \mathcal{F}_{c-attn}^s(\mathcal{F}_{c-attn}^t(\mathbf{H}^l, \mathbf{H}^P), \mathbf{H}^P)
\end{align}

where \( \mathbf{H}^P \) is the past motion features output by PME. \( \mathbf{H}^l \) is the input of \( l^{th} \) layer of FMD. \( \mathcal{F}_{c-attn}^s(\cdot) \) / \(\mathcal{F}_{c-attn}^t(\cdot) \) represents the spatial/temporal multi-head cross-attention calculation with the bottle neck. 

\subsection{Past Movement Guided Pretraining}
We apply the self-reconstruction pretraining paradigm to help the model learn the spatiotemporal representation of joints in motion sequences (as shown in the Fig.~\ref{fig:pipepine}), which compromise two proxy tasks: 1) past motion reconstruction; 2) future motion reconstruction. We will introduce each step in the following.

\paragraph{Past Motion Reconstruction} Given the past motion sequence \( \mathbf{X}^P \in \mathcal{R}^{T \times J \times K} \), we first calculate the motion velocity of each joint \( \mathbf{V} = \mathbf{X}_{2:T}^P - \mathbf{X}_{1:T-1}^P \in \mathcal{R}^{(T-1) \times J \times K} \). We find the threshold \( s \) in \( \mathbf{V} \) such that the \(r\)\% of the values in \( \mathbf{V} \) are greater than the threshold \( s \). The mask of the past motion sequence \( \mathbf{M}^P \) can be mathematically expressed as:

\begin{equation}
    \mathbf{M}^{P, (i, j)} =
    \left\{
    \begin{array}{ll}
        1 &, \mathbf{V}^{(i,j)} \geq s \\
        0 &, \mathbf{V}^{(i,j)} < s
    \end{array}
    \right.
\end{equation}

The masked past motion sequence can be computed by \( \mathbf{X}_{masked}^{P} = \mathbf{X}^P \times \mathbf{M}^P \). Then, the joint embedders are applied to obtain the masked motion embeds \( \mathbf{X}_{emb}^P \). Finally, we calculate the reconstructed motion sequence \( \mathbf{X}_{rec}^P = \mathcal{F}_{PH}(\mathcal{F}_{PME}(\mathbf{X}_{emb}^P)) \), where \( \mathcal{F}_{PH}(\cdot) \) and \( \mathcal{F}_{PME}(\cdot) \) are Predict Head and Past Motion Encoder. 

Through this past-motion self-reconstruction objective, the model learns to capture both the spatial structure within individual poses and the temporal dependencies across consecutive motions. This encourages the encoder to acquire rich and robust 3D human motion representations that benefit downstream tasks.

\paragraph{Future Motion Reconstruction}
Given the complete past motion sequence \( \mathbf{X}^P \), we first apply the joint embedder to obtain the motion embeddings \( \mathbf{X}_{emb}^P \), which are then encoded into past-motion features via \( \mathbf{H}^P = \mathcal{F}_{PME}(\mathbf{X}_{emb}^P) \). For the future motion sequence \( \mathbf{X}^F \in \mathbb{R}^{L \times J \times K} \), we adopt the same velocity-based masking strategy and embedding procedure, producing the masked embeddings \( \mathbf{X}_{emb}^F \). The masked future-motion embeddings \( \mathbf{X}_{emb}^F \) and the past-motion features \( \mathbf{H}^P \) are fed into the FMD module, where the PMG-Attention uses cross-attention to guide the reconstruction of masked future motion. The final reconstructed future motion is given by \( \mathbf{X}_{rec}^F = \mathcal{F}_{PH}(\mathcal{F}_{FMD}(\mathbf{X}_{emb}^F, \mathbf{H}^P)) \).

In this reconstruction task, the model completes masked future motions based on past observations. This encourages it to explicitly learn the fundamental dependency of future motion on past inputs, as well as the underlying priors of human movement patterns and motion trajectories, facilitating downstream tasks.

\subsection{Fine-tuning for Prediction and Understanding}
In the finetuning phase, we use the complete past motion sequence as input to the PME to obtain the past motion features (as shown in the Fig.\ref{fig:finetune}). In the FMD, we initialize the future query tokens with zeros (while preserving positional encodings) and decode the full future motion sequence guided by the past features. Given the past motion sequence $\mathbf{X}^P$, the prediction process is formulated as:
\begin{align}
    \mathbf{H}^P & = \mathcal{F}_{JE}(\mathcal{F}_{PME}(\mathbf{X}^P)) \\
    \mathbf{H}^F & = \mathcal{F}_{JE}(\mathcal{F}_{FMD}(\mathbf{H}_0^F, \mathbf{H}^P)) \\
    \mathbf{X}^F & = \mathcal{F}_{PH}(\mathbf{H}^F),
\end{align}
where $\mathbf{H}_0^F$ is the zero-initialized future state and $\mathcal{F}_{JE}$ denotes the joint embedder.

Beyond motion prediction, our framework also supports future motion understanding. Given the past-context representation $\mathbf{H}^P$, we apply a Future Semantic Decoder (FSD) $\mathcal{F}_{FSD}$ to generate a future semantic description:
\begin{align}
    \mathbf{Y}^F = \mathcal{F}_{FSD}(\mathbf{H}^F, \mathbf{H}^P),
\end{align}
where $\mathbf{Y}^F = [y_1, \ldots, y_M]$ is the predicted future text sequence. This enables the model to jointly perform future motion prediction and future motion understanding within a unified fine-tuning pipeline.

\subsection{Loss Function}

\paragraph{Loss Function of Pretraining Stage} Pretraining stage performs past sequence reconstruction and future sequence reconstruction. Given the past and future reconstructed results \( \hat{\mathbf{X}}_{rec}^P \) and \( \hat{\mathbf{X}}_{rec}^F \), and the corresponding ground-truth (i.e. complete sequence) \( \mathbf{X}_{rec}^P \) and \( \mathbf{X}_{rec}^F \), the loss function can be represented as follow:

\begin{equation}
    \begin{split}
    \mathcal{L}_{pretrain} 
    & = \frac{1}{TJ}\sum_{t=1}^{T}\sum_{j=1}^{J}||\mathbf{X}_{rec}^{P, (t,j)} - \hat{\mathbf{X}}_{rec}^{P, (t,j)} ||_{F}^2 \\
    & + \frac{\alpha}{LJ}\sum_{l=1}^{L}\sum_{j=1}^{J}||\mathbf{X}_{rec}^{F, (l,j)} - \hat{\mathbf{X}}_{rec}^{F, (l,j)} ||_{F}^2
    \end{split}
\end{equation}

where \( \alpha \) is the hyperparameter to balance the loss weight.

\paragraph{Loss Function of Finetune Stage}
In the finetuning stage, the overall objective integrates motion reconstruction and semantic prediction into a single loss:

\begin{equation}
\begin{aligned}
\mathcal{L}_{finetune}
&=
\underbrace{
\frac{1}{LJ}
\sum_{l=1}^{L}\sum_{j=1}^{J}
\left\| \mathbf{X}_{j}^{l} - \hat{\mathbf{X}}_{j}^{l} \right\|_{F}^{2}
}_{\mathcal{L}_{motion}}
\\[4pt]
&\quad
+
\lambda
\underbrace{
\left(
- \sum_{d=1}^{D} 
\log p(\mathbf{Y}^{F}_{d} \mid \hat{\mathbf{Y}}^{F})
\right)
}_{\mathcal{L}_{text}}.
\end{aligned}
\end{equation}

Jointly optimizing the two losses encourages mutually reinforcing representations, leading to more accurate future motion prediction and more consistent semantic generation.

\begin{table*}[t!]
\centering
\setlength\tabcolsep{5pt}
% \scriptsize
\small
\caption{Comparisons of MPJPEs between our proposed method with other \textit{state-of-the-art} methods for short-term prediction on Human3.6. We show the results of detailed actions such as walking and eating. The best results are highlighted in bold, and the second best results is underlined. }
\begin{tabular}{ c | c c c c | c c c c | c c c c | c c c c } 
\toprule

Motion & \multicolumn{4}{c|}{Walking} & \multicolumn{4}{c|}{Eating} & \multicolumn{4}{c|}{Smoking} & \multicolumn{4}{c}{Discussion} \\
\hline
Milliseconds & 80 & 160 & 320 & 400 & 80 & 160 & 320 & 400 & 80 & 160 & 320 & 400 & 80 & 160 & 320 & 400 \\ 
\hline
DSTD-GC~\cite{fu2023learning} & 11.1 & 22.4 & 38.8 & 45.2 & 7.0 & 15.5 & 31.7 & 39.2 & 6.6 & 14.8 & 29.8 & 36.7 & 10.0 & 24.4 & 54.5 & 67.4 \\
AuxFormer~\cite{xu2023auxiliary} & \underline{8.9} & \underline{16.9} & \textbf{30.1} & \textbf{36.1} & \underline{6.4} & \underline{14.0} & \textbf{28.8} & \textbf{35.9} & \underline{5.7} & \underline{11.4} & \underline{22.1} & \underline{27.9} & \underline{8.6} & \underline{18.8} & \underline{38.8} & \underline{49.2} \\
CIST-GCN~\cite{medina2024context} & 11.8 & 23.4 & 40.5 & 46.5 & 6.7 & 14.8 & 29.8 & \underline{36.8} & 7.3 & 15.6 & 31.0 & 38.0 & 10.2 & 23.7 & 52.3 & 65.3 \\

Ours & \textbf{8.6} & \textbf{16.5} & \underline{31.2} & \underline{37.7} & \textbf{6.1} & \textbf{13.7} & \underline{29.3} & \underline{36.8} & \textbf{5.3} & \textbf{10.3} & \textbf{19.7} & \textbf{25.2} & \textbf{7.7} & \textbf{16.3} & \textbf{31.7} & \textbf{40.9} \\
\hline

\hline
Motion & \multicolumn{4}{c|}{Direction} & \multicolumn{4}{c|}{Greeting} & \multicolumn{4}{c|}{Phoning} & \multicolumn{4}{c}{Posing} \\
\hline
Milliseconds & 80 & 160 & 320 & 400 & 80 & 160 & 320 & 400 & 80 & 160 & 320 & 400 & 80 & 160 & 320 & 400 \\ 
\hline
DSTD-GC~\cite{fu2023learning} & 6.9 & 17.4 & \underline{40.1} & 51.7 & 14.3 & 33.5 & 72.2 & 87.3 & 8.5 & 19.2 & 40.3 & 49.9 & 10.1 & 25.4 & 60.6 & 77.3 \\
AuxFormer~\cite{xu2023auxiliary} & \underline{6.8} & \underline{17.0} & 40.3 & 51.6 & \underline{13.5} & 31.3 & 69.2 & 85.4 & \underline{7.9} & \underline{17.3} & \textbf{37.4} & \textbf{47.2} & \underline{8.8} & \underline{19.1} & \underline{39.2} & \underline{51.0} \\
CIST-GCN~\cite{medina2024context} & 7.3 & 18.1 & 43.6 & 55.3 & 13.7 & \underline{31.0} & \underline{65.7} & \underline{79.9} & 8.6 & 18.5 & 39.3 & 49.6 & 9.6 & 23.7 & 57.7 & 75.0 \\

Ours & \textbf{6.3} & \textbf{15.8} & \textbf{39.8} & \underline{51.4} & \textbf{10.8} & \textbf{26.4} & \textbf{62.7} & \textbf{79.8} & \textbf{7.5} & \textbf{17.1} & \underline{37.9} & \underline{48.0} & \textbf{8.2} & \textbf{16.9} & \textbf{32.8} & \textbf{43.5} \\

\bottomrule
\end{tabular}
\label{h36m-short-all}
\vspace{-0.25cm}
\end{table*}

\begin{table}[t]
\centering
\small
\caption{Comparisons of MPJPEs between our proposed method with other \textit{state-of-the-art} methods for short- and long-term prediction on Human3.6M. We show the average result of all actions. The best results are highlighted in bold.}
\resizebox{\linewidth}{!}{
\begin{tabular}{ c | c c c c c c } 
\toprule

Motion & \multicolumn{6}{c}{Average} \\
\hline
Milliseconds & 80 & 160 & 320 & 400 & 560 & 1000 \\ 
\hline
DMGNN~\cite{li2020dynamic} & 17.0 & 33.6 & 65.9 & 79.7 & 103.0 & 137.2 \\
MSR~\cite{dang2021msr} & 12.1 & 25.6 & 51.6 & 62.9 & 81.1 & 114.2 \\
Mixer~\cite{bouazizi2022motionmixer} & 11.1 & 24.0 & 51.5 & 64.0 & 83.5 & 116.5 \\
PGBIG~\cite{ma2022progressively} & 10.3 & 22.7 & 47.4 & 58.5 & 76.9 & 110.3 \\
SPGSN~\cite{li2022skeleton} & 10.4 & 22.3 & 47.1 & 58.3 & 77.1 & 108.8 \\
siMLPe~\cite{guo2023back} & 13.4 & 23.7 & 47.4 & 58.0 & 77.9 & 111.7 \\
DSTD-GC~\cite{fu2023learning} & 10.4 & 23.3 & 48.8 & 59.8 & 77.8 & 111.0 \\
AuxFormer~\cite{xu2023auxiliary} & 9.5 & 20.6 & 43.4 & 54.1 & 75.3 & 107.0 \\
CIST-GCN~\cite{medina2024context} & 10.5 & 23.2 & 47.9 & 59.0 & 77.2 & 110.3 \\
GCNext~\cite{wang2024gcnext} & 9.3 & 21.5 & 45.5 & 56.4 & 74.7 & 108.7 \\
ALIEN~\cite{wei2025alien} & 9.7 & 21.8 & 47.0 & 58.3 & - & - \\
\hline
Ours & \textbf{8.8} & \textbf{19.3} & \textbf{41.1} & \textbf{51.8} & \textbf{70.0} & \textbf{101.5} \\
\bottomrule
\end{tabular}
}
\label{h36m-short-long-avg}
\vspace{-0.5cm}
\end{table}

\section{Experiments}
\label{sec:experiments}

\subsection{Dataset}

\paragraph{Human3.6M} Human3.6M~\cite{ionescu2013human3} is a large-scale, accurately annotated 3D human pose dataset that captures the motions of 11 professional actors in daily life scenarios using motion capture technology, totaling over 3.6 million pose instances. Following previous works~\cite{ma2022progressively, li2022skeleton},  only 22 joints are used, focusing on those that exhibit significant movement, and all sequences are down-sampled to 25 fps.

\paragraph{The 3D Poses in the Wild (3DPW)} 3DPW~\cite{von2018recovering} stands out as a pioneering dataset in motion related tasks, primarily due to its focus on capturing human motion in natural settings outside of the typical laboratory environment. This dataset provides a rich collection of video sequences captured at a frame rate of 30 frames per second, which includes diverse activities such as `Walking', `Running', and interacting with various objects.

\paragraph{The Archive of Motion Capture as Surface Shapes (AMASS)} AMASS~\cite{mahmood2019amass} consolidates 15 existing motion capture datasets into a single, standardized framework. By converting these diverse datasets into a common representation, AMASS provides over 40 hours of meticulously annotated motion data. Following the previous works~\cite{shi2023multi, wang2024dynamic}, we discarded the detailed joints of the hand and sampled the sequence to 24fps.

\paragraph{FineMotion} To incorporate fine-grained action semantics, we adopt FineMotion\footnote{\url{https://github.com/CVI-SZU/FineMotion}}, an open-source extension of HumanML3D that re-labels all motions with detailed body-part movement descriptions. FineMotion segments each sequence into 0.5-second snippets and provides both human-annotated and automatically generated descriptions. Based on this resource, we construct our prediction dataset using a 2-second sliding window (0.5s past motion and 1.5s future motion). For each window, we concatenate the snippet-level descriptions corresponding to the future segment to form its semantic target. This process yields 12,628 human-annotated samples and 248,126 automatically annotated samples, which we split into train/val/test sets following an 80\% / 5\% / 15\% ratio.

\begin{figure*}[t!]
\centering
\includegraphics[width=0.95\textwidth]{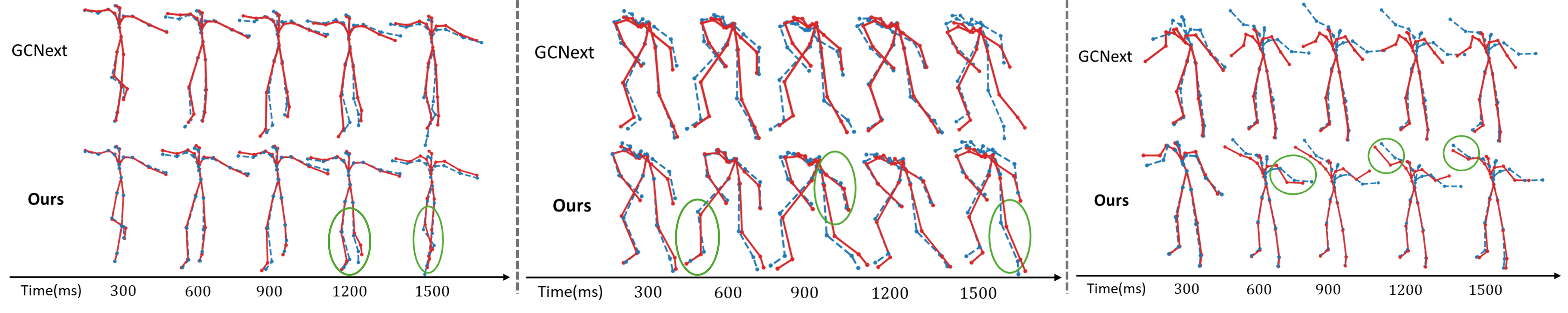}
\caption{Visualization results of motion prediction. The red lines represent predicted motions, and the blue lines represent ground-truths.}
\label{fig:vis-pred}
\end{figure*}

\begin{figure*}[t!]
\centering
\includegraphics[width=0.95\textwidth]{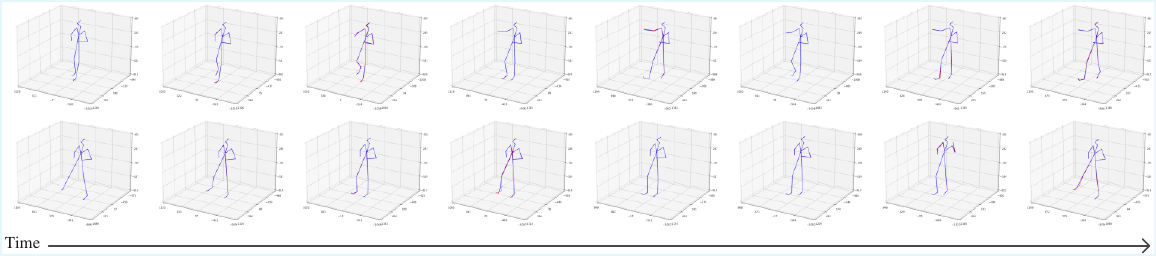}
\caption{Visualization results of motion reconstruction on H36M. The blue lines represent reconstructed motions, and the red lines represent ground-truths.}
\label{fig:vis-recon}
\vspace{-0.5cm}
\end{figure*}

\begin{figure}[t!]
\centering
\includegraphics[width=0.45\textwidth]{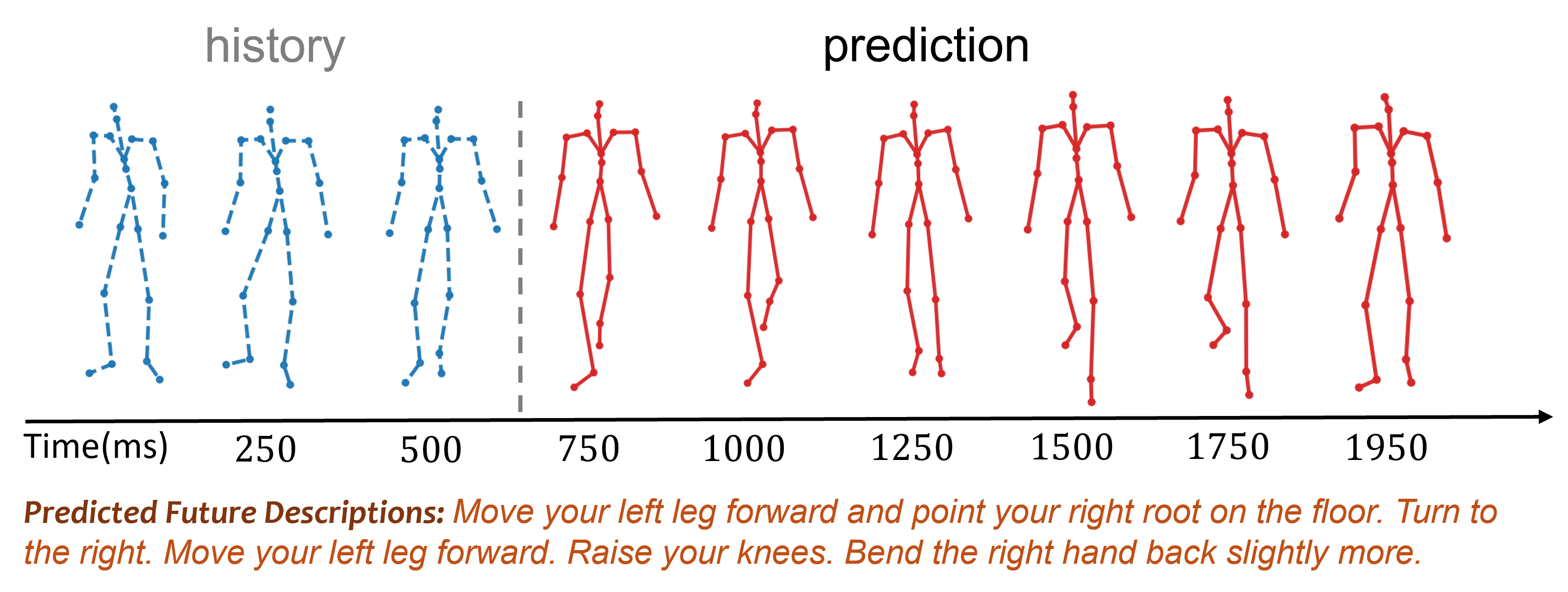}
\caption{Visualization of motion prediction and understanding.}
\label{fig:vis-pred-und}
\vspace{-0.5cm}
\end{figure}

\subsection{Experimental Settings} For the PME, FMD and FSU, the number of modules are all set to 3. The number of feature channels remains consistent at 128 throughout the PME and FMD, and remains at 256 throughout the FSU. In the attention calculation, we employed 8 heads, with each head comprising 32 feature channels. During the pretraining phase, the loss weight (denoted as \( \alpha \)) is set to 1, indicating an equal contribution of loss from both past and future sequence reconstruction. During the finetune phase, the loss weight (denoted as \(\lambda\)) is set to 0.1. The parameters in FSU are updated every 5 epoch. 

\subsection{Comparison to State-of-the-art Methods}

We compare our method with the recent \textit{state-of-the-art} methods~\cite{li2020dynamic, mao2021multi, aksan2021spatio, dang2021msr, bouazizi2022motionmixer, ma2022progressively, li2022skeleton, guo2023back, fu2023learning, xu2023auxiliary, wang2024dynamic, medina2024context, wang2024gcnext, wei2025alien}. To ensure fair comparisons, we re-evaluated some methods on the entire test set to align the testing standard using their official pretrained models, or retrain the model (Only for the cases where the official pre-training model is not provided) using their default hyperparameters on the Human3.6M.

\paragraph{Comparisons on Human3.6M} 
We use MPJPE (Mean Per Joint Position Error) as the evaluation metric, which is widely used in motion prediction tasks~\cite{ma2022progressively, li2022skeleton, xu2023auxiliary}. \cref{h36m-short-all} shows the quantitative comparisons of short-term motion prediction on Human3.6M dataset, which aims to predict the poses within 400 milliseconds. Detailed actions are listed. As can be seen, our method outperforms the others on most actions, especially the ``Smoking'', ``Greeting'', and ``Posing''. \cref{h36m-short-long-avg} displays the average results of short- and long-term predictions for all actions on Human3.6M. Our approach effectively reduces prediction errors by 5.4\%/10.2\%/9.7\%/8.2\%/6.3\%/6.6\% at 80ms/160ms/320ms/400ms/560ms/1000ms.

\begin{table}[h!]
\centering
\small
% \scriptsize
\setlength\tabcolsep{2.5pt}
\caption{Comparisons of MPJPEs between our proposed method with other SOTA methods for both short-term and long-term prediction on AMASS. The best results are highlighted in bold.}
% \resizebox{\textwidth}{15mm}{
\resizebox{\linewidth}{!}{
\begin{tabular}{ c c c c c c c c c } 
\toprule[1.25pt]
Milliseconds & 80 & 160 & 320 & 400 & 560 & 720 & 880 & 1000 \\ 
\hline
LTD-10-10~\cite{mao2019learning} & 10.3 & 19.3 & 36.6 & 44.6 & 61.5 & 75.9 & 86.2 & 91.2 \\
LTD-10-25~\cite{mao2019learning} & 11.0 & 20.7 & 37.8 & 45.3 & 57.2 & 65.7 & 71.3 & 75.2 \\
HisRep~\cite{mao2020history} & 11.3 & 20.7 & 35.7 & 42.0 & 51.7 & 58.6 & 63.4 & 67.2 \\
Mao W, et al.~\cite{mao2021multi} &  11.0 & 20.3 & 35.0 & 41.2 & 50.7 & 57.4 & 61.9 & 65.8 \\
siMLPe~\cite{guo2023back} & 10.8 & 19.6 & 34.3 & 40.5 & 50.5 & 57.3 & 62.4 & 65.7 \\
DD-GCN~\cite{wang2024dynamic} & 10.4 & 19.1 & 33.6 & 39.8 & 49.3 & 56.5 & 61.3 & 64.6 \\

\hline
Ours & \textbf{8.6} & \textbf{15.3} & \textbf{27.7} & \textbf{33.2} & \textbf{42.0} & \textbf{48.3} & \textbf{53.0} & \textbf{56.0} \\

\bottomrule[1.25pt]
\end{tabular}
}
\label{amass}
\end{table}

 \paragraph{Comparisons on AMASS}
 \cref{amass} shows the comparisons on AMASS dataset. Our method achieves a reduction in prediction errors of 15.9\%. We found that our method exhibits the most significant improvement on this dataset. This can be attributed to the larger data size of AMASS compared to Human3.6M and 3DPW. The self-reconstruction pretraining enables our model to learn a more comprehensive representation of motion on such a large dataset.

\begin{table}[h!]
\centering
\setlength\tabcolsep{2pt}
% \scriptsize
\small
\caption{Comparisons of MPJPEs between our proposed method with other \textit{state-of-the-art} methods for short- and long-term prediction on 3DPW. The best results are highlighted in bold.}
\resizebox{\linewidth}{!}{
\begin{tabular}{ c | c c c c c c } 
\toprule

  & \multicolumn{6}{c}{Average MPJPE}  \\
\hline
Milliseconds & 100 & 200 & 400 & 600 & 800 & 1000 \\ 
\hline
DMGNN~\cite{li2020dynamic} & 17.80 & 37.11 & 70.38 & 94.12 & 109.67 & 123.93 \\
HisRep~\cite{mao2020history} & 15.88 & 35.14 & 66.82 & 93.55 & 107.63 & 114.75 \\
MSR-GCN~\cite{dang2021msr} & 15.70 & 33.48 & 65.02 & 93.81 & 108.15 & 116.31 \\
PGBIG~\cite{ma2022progressively} & 17.66 & 35.32 & 67.83 & 89.60 & 102.59 & 109.41 \\
SPGSN~\cite{li2022skeleton} & 15.39 & 32.91 & 64.54 & 91.62 & 103.98 & 111.05 \\
AuxFormer~\cite{xu2023auxiliary} & 14.21 & 30.04 & 58.50 & 89.45 & 100.78 & 107.45 \\
ALIEN~\cite{wei2025alien} & 14.83 & 32.59 & 62.06 & 90.25 & 103.62 & - \\

\hline
Ours & \textbf{13.85} & \textbf{29.11} & \textbf{57.06} & \textbf{87.35} & \textbf{97.93} & \textbf{103.37} \\

\bottomrule
\end{tabular}
}
\label{3dpw}
\vspace{-0.5cm}
\end{table}

\paragraph{Comparisons on 3DPW}
The 3DPW dataset comprises data collected from outside the laboratory setting, which are widely used to verify the power of the model in real scenarios. As shown in \cref{3dpw}, our approach performed best across all frames, showcasing its robustness in handling complex interactive scenarios.

\paragraph{Comparisons on Joint Prediction and Understading}
As shown in \cref{tab:finemotion-prediction}, our method consistently achieves lower MPJPEs on the processed FineMotion dataset across all prediction horizons compared to prior approaches. Notably, introducing the understanding branch further improves motion prediction quality, indicating that capturing semantic cues of future movements can, in turn, benefit joint-level forecasting.

As shown in \cref{tab:finemotion-understanding}, the IDs ``A'' and ``B'' refer to models trained on auto-labeled and human-labeled text annotations, respectively. Our lightweight semantic prediction head achieves competitive or superior performance across several key metrics when compared with MGMotionLLM~\cite{wu2025mg}, despite the latter being a large-scale language model designed for motion–text generation. Moreover, when the motion prediction loss is removed, the performance on future text generation also degrades, indicating that accurate motion forecasting provides essential guidance for understanding future semantic.

\begin{table}[t]
\centering
\caption{Comparison of classification accuracy and clustering quality on Human3.6M using network-output features. CLS accuracy evaluates discriminativeness, while inter- and intra-class distances measure feature separability.}
\resizebox{\linewidth}{!}{
\begin{tabular}{l|l|ccc}
\toprule
Task                     & Metrics     & w/o Network & AuxFormer~\cite{xu2023auxiliary} & Ours   \\ \hline
CLS                      & Accuracy\(\uparrow\)    & 25.6\%      & 80.8\%    & \textbf{95.0\%} \\ \hline
\multirow{2}{*}{Cluster} & Inter dist.\(\uparrow\) & --          & 59.7      & \textbf{64.4}   \\
                         & Intra dist.\(\downarrow\) & --          & 145.6     & \textbf{135.7}  \\
\bottomrule
\end{tabular}
}
\label{representation}
\vspace{-0.25cm}
\end{table}

\paragraph{Representation Capability}
We evaluate the quality of the learned motion representation by extracting penultimate-layer features and training a lightweight linear classifier on them. As shown in Table~\ref{representation}, our method achieves substantially higher classification accuracy than both w/o Network and AuxFormer~\cite{xu2023auxiliary}, demonstrating stronger feature discriminativeness. In addition, our features present larger inter-class distances and smaller intra-class distances, indicating better clustering behavior and clearer semantic separability. For the clustering metrics, we randomly sample 100 instances from each of the 15 action categories in Human3.6M to compute the intra- and inter-class distances.

\subsection{Visualization of Prediction and Reconstruction}
\cref{fig:vis-pred} presents the visualization of prediction results. We conducted comparisons with GCNext~\cite{wang2024gcnext} on challenging samples. Our model successfully captures significant movements of the arms and legs. This can be attributed to the utilization of past motion guidance, which enables the model to comprehend the influence of previous motion patterns on future sequences. \cref{fig:vis-pred-und} showcases the joint results of future motion prediction and future motion understanding, where our method consistently outputs motion sequences and semantic descriptions that are highly aligned. 
Furthermore, We provide reconstruction results with a mask rate of 75\% in \cref{fig:vis-recon}. Similar to the property of mask-based pretraining~\cite{he2022masked}, even with a high mask rate of 75\% for joints, the model can still accurately reconstruct the location of them.

\begin{table}[t]
\centering
\setlength\tabcolsep{5pt}
\small
\caption{Comparisons of MPJPEs between our proposed method with other \textit{state-of-the-art} methods on the processed FineMotion dataset. The best results are highlighted in bold.}
\vspace{-0.25cm}
\resizebox{\linewidth}{!}{
\begin{tabular}{ c | c c c c c } 
\toprule
  & \multicolumn{5}{c}{Average MPJPE} \\
\hline
Milliseconds & 300 & 600 & 900 & 1200 & 1500 \\
\hline
AuxFormer~\cite{xu2023auxiliary} & 36.07 & 62.31 & 76.95 & 86.11 & 94.22 \\
GCNext~\cite{wang2024gcnext}    & 44.05 & 73.19 & 88.14 & 97.03 & 103.9 \\
\hline
Ours w/o Und. & 34.96 & 57.39 & 69.08 & 76.58 & 84.95 \\
Ours                             & \textbf{34.69} & \textbf{56.71} & \textbf{67.99} & \textbf{75.33} & \textbf{83.65} \\
\bottomrule
\end{tabular}
}
\label{tab:finemotion-prediction}
\vspace{-0.25cm}
\end{table}

\begin{table}[t]
\centering
\setlength\tabcolsep{5pt}
\small
\caption{Comparisons of future motion understanding between our proposed method with MGMotionLLM~\cite{wu2025mg} on the processed FineMotion dataset. The best results are highlighted in bold.}
\vspace{-0.25cm}
\resizebox{\linewidth}{!}{
\begin{tabular}{ c | c | c c c c } 
\toprule
ID & Method & BLEU@4 & BLEU@1 & ROUGE & CIDEr \\
\hline
\multirow{3}{*}{A} 
& MGMotionLLM~\cite{wu2025mg} & \textbf{25.3} & \textbf{15.8} & \textbf{38.0} & 12.5 \\
& Ours w/o Prediction & 15.6 & 7.9 & 31.2 & 36.2 \\
& Ours                       & 17.1 & 9.4 & 33.5 & \textbf{47.6} \\
\hline
\multirow{3}{*}{B}
& MGMotionLLM~\cite{wu2025mg} & 13.1 & 7.5 & \textbf{35.8} & 3.3 \\
& Ours w/o Prediction & 15.0 & 7.1 & 31.7 & \textbf{25.8} \\
& Ours                       & \textbf{18.2} & \textbf{7.8} & 32.5 & 24.2 \\
\bottomrule
\end{tabular}
}
\label{tab:finemotion-understanding}
\vspace{-0.25cm}
\end{table}

\section{Ablation Study}
\label{sec:ablation_study}

We explore the impact of components of our network, including the mask strategy, pretraining paradigm. We provide studies on model architecture in the supplementary materials. All experiments were carried out on the Human3.6M dataset.

\paragraph{Mask Strategy}
\cref{ablation_study} presents the results of two cases: A, without utilizing the velocity-based mask strategy (i.e. randomly mask 75\% of the joints), and F, which represents the complete model. It can be seen that using the velocity-based mask reduced the average prediction errors by 2.3\%. \cref{ablation_study} also shows the comparisons of different mask rates (ranging from 25\% to 75\%), represented by B, C, F. The high mask rate facilitates the model to learn the space-time representation during the self-reconstruction process.

\paragraph{Effects of Pretraining Paradigm}
We show the results of pretraining the model by denoising in \cref{ablation_study} E. Compared with the self-reconstruction, denoising pretraining method lacks the exploration of the spatiotemporal relationship between joints. The result without pretraining is denoted as D, which clearly demonstrates the crucial role of our pretraining strategy in improving prediction performance.

\begin{table}[t]
\centering
\setlength\tabcolsep{4.7pt}
% \scriptsize
\small
\caption{Impact of mask strategy and pretraining paradigm. \(r\) is the mask rate. ``denoise'' means we conduct self-supervised pretraining by denoising. }
\vspace{-0.25cm}
\begin{tabular}{ c | c | c c c c } 
\toprule

% \diagbox{description}{ms}

ID &  & 80ms & 160ms & 320ms & 400ms  \\ 
\hline
A & random mask & 9.0 & 19.7 & 42.4 & 52.8 \\
\hline
B & \( r = 25 \% \) & 9.3 & 20.2 & 42.9 & 53.7  \\
C & \( r = 50 \% \) & 9.1 & 20.0 & 42.3 & 52.6  \\
\hline
D & w/o pretraining & 9.6 & 21.9 & 46.2 & 56.8  \\ 
E & denoise & 9.2 & 20.4 & 42.9 & 53.8  \\ 
\hline
F & complete (\( r = 75 \% \)) & \textbf{8.8} & \textbf{19.3} & \textbf{41.1} & \textbf{51.8}  \\

\bottomrule
\end{tabular}
\label{ablation_study}
\vspace{-0.5cm}
\end{table}

\section{Conclusion}
\label{sec:conclusion}
In this work, we addressed the core limitations of end-to-end regression in motion prediction by introducing a two-stage framework that separates representation learning from prediction. Through past-guided future reconstruction and velocity-based masking, our self-supervised pretraining effectively captures intrinsic spatiotemporal dependencies. The joint fine-tuning of motion prediction and semantic future understanding further enhances coherence and realism.

{
    \small
    \bibliographystyle{ieeenat_fullname}
    \bibliography{main}
}

\clearpage
\setcounter{page}{1}
\maketitlesupplementary

\begin{figure*}[t]
\centering
\includegraphics[width=0.90\textwidth]{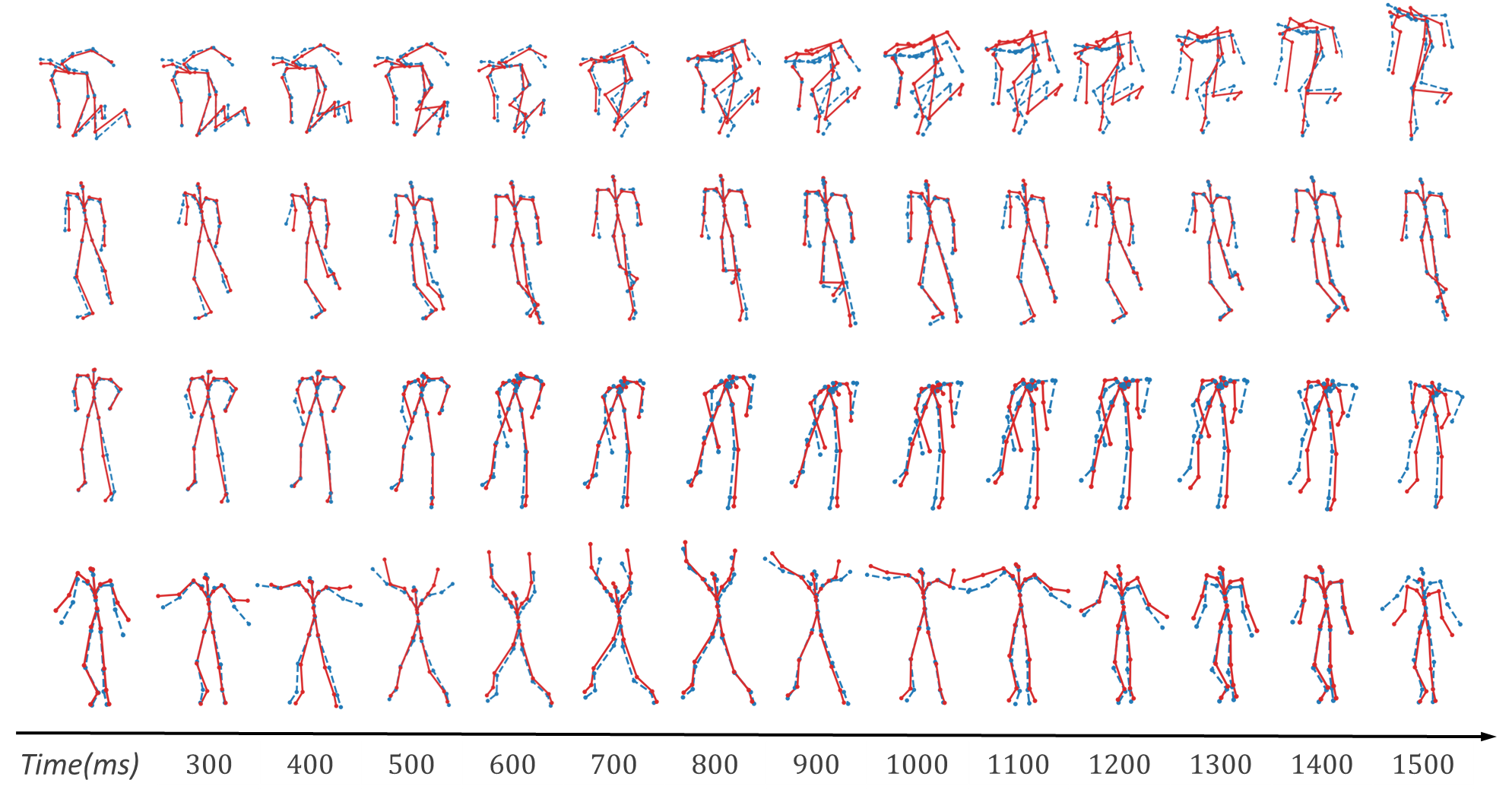}
\caption{Visualization of motion prediction results on the processed FineMotion dataset.}
\label{suppl-vis-finemotion}
\end{figure*}

\begin{figure}[t]
\centering
\includegraphics[width=0.48\textwidth]{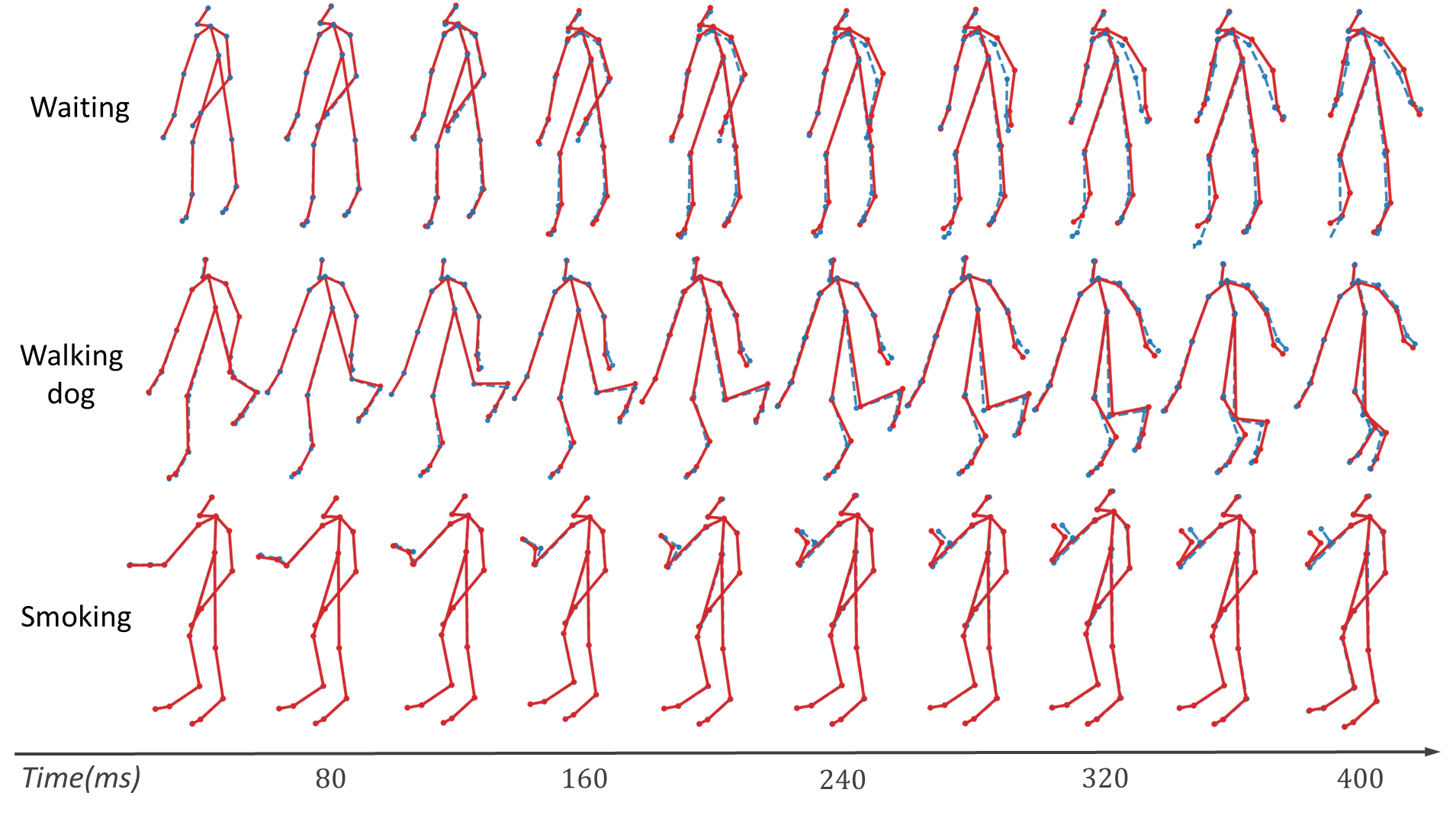}
\caption{Visualization of motion prediction results (``Posing'', ``Walking'', ``Greeting'') on Human3.6M dataset.}
\label{suppl-vis-h36m-2}
\end{figure}

\begin{figure}[t]
\centering
\includegraphics[width=0.48\textwidth]{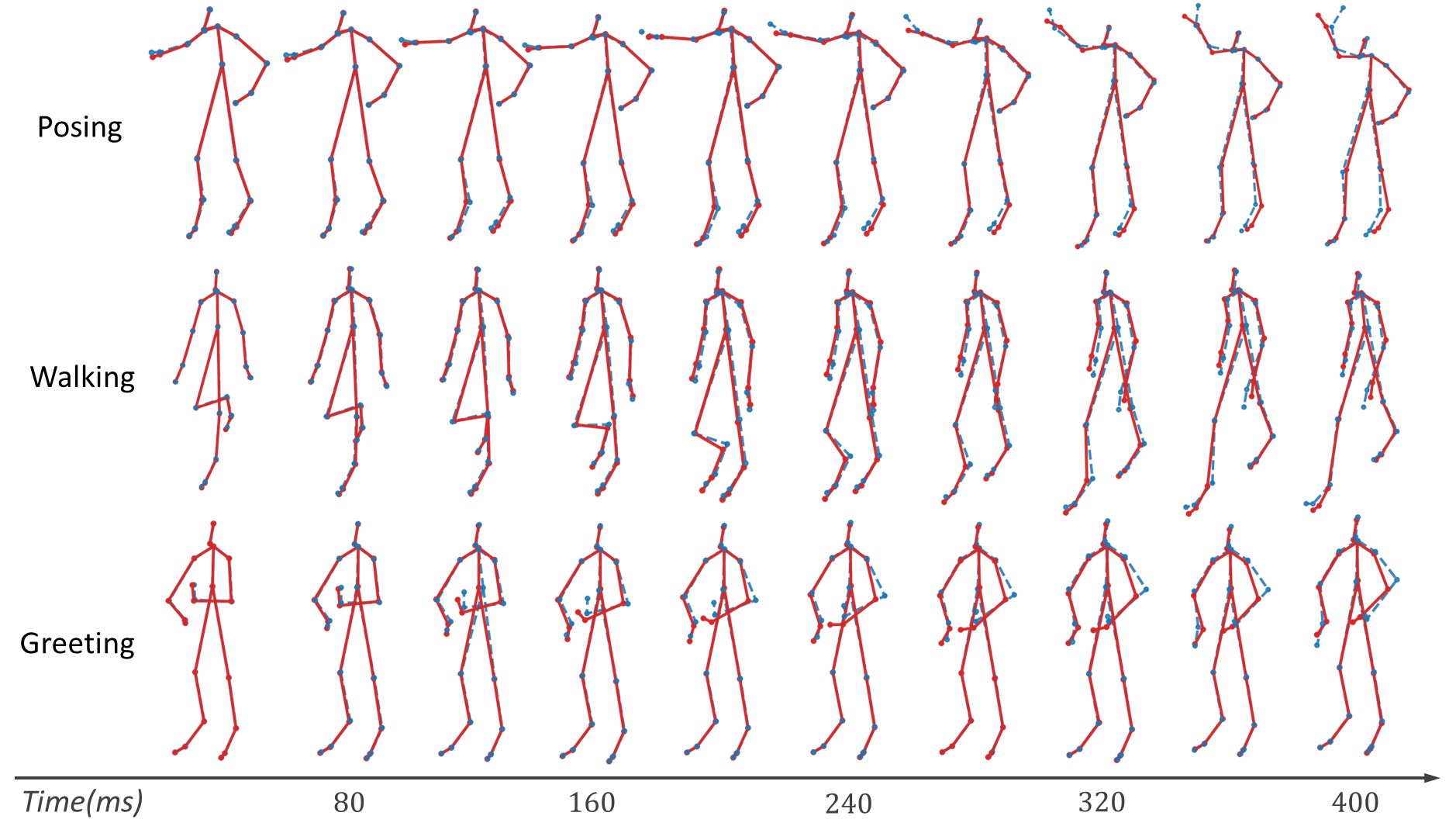}
\caption{Visualization of motion prediction results (``Posing'', ``Walking'', ``Greeting'') on Human3.6M dataset.}
\label{suppl-vis-h36m-1}
\end{figure}

In this supplementary material, we provide additional technical details, extended results, and further qualitative analyses that complement the main paper. Specifically, we include a detailed explanation of the spatiotemporal attention mechanisms used in our architecture (\cref{sec:suppl-model}), implementation details and training configurations (\cref{sec:suppl-details}), full motion prediction results on the Human3.6M dataset (\cref{sec:suppl-result-h36m}), comprehensive ablation studies on model components and architectural choices (\cref{sec:suppl-ablation}), and extensive visualizations, including motion prediction results, and attention feature maps (\cref{sec:suppl-vis-results} and \cref{sec:suppl-vis-attn}). We also discuss the mask strategy and reconstruction performance in \cref{sec:suppl-mask}. Together, these supplementary materials offer a deeper understanding of our framework and provide further evidence of its effectiveness and robustness.

\section{Detailed Explanation of Model Architecture}
\label{sec:suppl-model}

We provide additional details of the two attention modules used in our framework: the SpatioTemporal Attention (ST-Attention) in the Past Motion Encoder (PME), and the spatiotemporal cross-attention in the Future Motion Decoder (FMD). Both modules follow the standard Transformer design, where \emph{each} attention sub-layer (temporal or spatial) is immediately followed by a position-wise feed-forward network (FFN), each equipped with its own residual connection and layer normalization.

\paragraph{ST-Attention (within PME).}
Each ST-Attention layer contains a temporal attention block and a spatial attention block, applied sequentially. Given the input to the \( l^{th} \) layer \( \mathbf{H}^{l} \), the temporal self-attention is computed as:
\begin{equation}
\mathbf{Q}^{t}, \mathbf{K}^{t}, \mathbf{V}^{t} = \mathcal{F}^{t}_{QKV}(\mathbf{H}^{l}),
\end{equation}
\begin{equation}
\mathbf{A}^{t}
= \operatorname{softmax}\!\left(
\frac{\mathbf{Q}^{t}(\mathbf{K}^{t})^{\mathrm{T}}}{\sqrt{d_k}}
\right)\mathbf{V}^{t}.
\end{equation}
A position-wise FFN follows:
\begin{equation}
\mathbf{H}^{l,t}
= \mathrm{LN}\!\left(\mathbf{A}^{t} + \mathrm{FFN}(\mathbf{A}^{t})\right).
\end{equation}

The refined features \( \mathbf{H}^{l,t} \) then pass through a spatial attention block:
\begin{equation}
\mathbf{Q}^{s}, \mathbf{K}^{s}, \mathbf{V}^{s}
= \mathcal{F}^{s}_{QKV}(\mathbf{H}^{l,t}),
\end{equation}
\begin{equation}
\mathbf{A}^{s}
= \operatorname{softmax}\!\left(
\frac{\mathbf{Q}^{s}(\mathbf{K}^{s})^{\mathrm{T}}}{\sqrt{d_k}}
\right)\mathbf{V}^{s},
\end{equation}
followed again by an FFN:
\begin{equation}
\mathbf{H}^{l+1}
= \mathrm{LN}\!\left(\mathbf{A}^{s} + \mathrm{FFN}(\mathbf{A}^{s})\right).
\end{equation}

\paragraph{Spatiotemporal Cross-Attention (within FMD).}
FMD incorporates the encoded past motion features \( \mathbf{H}^{P} \) via spatiotemporal cross-attention, where queries come from the current decoding state and keys/values come from the past sequence. As in PME, each temporal and spatial block has an independent FFN.

Temporal cross-attention is given by:
\begin{equation}
\mathbf{Q}^{t} = \mathcal{F}^{t}_{Q}(\mathbf{H}^{l}), \quad
\mathbf{K}^{t},\mathbf{V}^{t} = \mathcal{F}^{t}_{KV}(\mathbf{H}^{P}),
\end{equation}
\begin{equation}
\mathbf{A}^{t}
= \operatorname{softmax}\!\left(
\frac{\mathbf{Q}^{t}(\mathbf{K}^{t})^{\mathrm{T}}}{\sqrt{d_k}}
\right)\mathbf{V}^{t},
\end{equation}
followed by:
\begin{equation}
\mathbf{H}^{l,t}
= \mathrm{LN}\!\left(\mathbf{A}^{t} + \mathrm{FFN}(\mathbf{A}^{t})\right).
\end{equation}

Spatial cross-attention further injects structural cues from past motion:
\begin{equation}
\mathbf{Q}^{s} = \mathcal{F}^{s}_{Q}(\mathbf{H}^{l,t}), \quad
\mathbf{K}^{s},\mathbf{V}^{s} = \mathcal{F}^{s}_{KV}(\mathbf{H}^{P}),
\end{equation}
\begin{equation}
\mathbf{A}^{s}
= \operatorname{softmax}\!\left(
\frac{\mathbf{Q}^{s}(\mathbf{K}^{s})^{\mathrm{T}}}{\sqrt{d_k}}
\right)\mathbf{V}^{s},
\end{equation}
with its FFN:
\begin{equation}
\mathbf{H}^{l+1}
= \mathrm{LN}\!\left(\mathbf{A}^{s} + \mathrm{FFN}(\mathbf{A}^{s})\right).
\end{equation}

This spatiotemporal cross-attention design enables FMD to effectively leverage both temporal evolution and spatial structure in the past sequence when generating future motions.

\section{Implementation Details}
\label{sec:suppl-details}
We implement all models using PyTorch 2.0 and train them on a single NVIDIA RTX4090 GPU. The batch size is set to 24, and the Adam optimizer is used with an initial learning rate of 0.0005. A cosine annealing schedule is adopted to gradually decay the learning rate, and the overall GPU memory consumption remains below 6GB, ensuring that our method can be easily deployed on a wide range of hardware platforms.

For network configuration, the Past Motion Encoder (PME), Future Motion Decoder (FMD), and Future Semantic Unit (FSU) each contain 3 stacked modules. The feature dimension is kept constant at 128 within PME and FMD, and at 256 within FSU. We employ 8-head attention, with each head containing 32 channels. During pretraining, the reconstruction losses of past and future motions share equal weights (=1). In the finetuning stage, the loss weight of text generation is set to 0.1, and the parameters of the FSU are updated every 5 epochs.

For self-supervised training on H36M, 3DPW, and AMASS, we train the model for 40 epochs and then finetune it for an additional 20 epochs on the motion prediction task. The best checkpoint on the validation set is selected after the self-supervised stage. For the processed FineMotion dataset, we train for 300 epochs in the self-supervised phase and subsequently finetune for 200 epochs on motion prediction.

\section{More visualization results}
\label{sec:suppl-vis-results}
We provide additional visualizations of motion prediction results on the processed FineMotion dataset in \cref{suppl-vis-finemotion}, and further show qualitative results on the Human3.6M dataset in \cref{suppl-vis-h36m-1} and \cref{suppl-vis-h36m-2}.
On Human3.6M, our method accurately predicts joint movements and maintains consistent body configurations over time. In particular, fine-grained actions involving limb articulations, such as arm swings, remain temporally coherent and physically plausible.

Compared with Human3.6M, the motions in FineMotion are more complex and the prediction horizon is significantly longer (up to 1.5 seconds). Even under these challenging conditions, our model successfully captures the correct motion tendencies and preserves the overall dynamics of large-amplitude arm movements and posture transitions. This robustness benefits from our self-supervised pretraining strategy, which encourages the model to learn implicit spatiotemporal constraints among joints. Nevertheless, we observe noticeable errors at distal joints, especially in highly articulated or fast-changing motion segments. This limitation partly arises from the inherent uncertainty and variability in the execution of fine-grained actions. In future work, we plan to address this issue by explicitly modeling motion uncertainty and incorporating stronger motion priors into the prediction framework.

\section{Ablation Study on Model Architecture}
\label{sec:suppl-ablation}

\begin{table}[t]
\centering
\setlength\tabcolsep{4.0pt}
\caption{Ablation study on future text generation performance (future motion semantic understanding).}
\begin{tabular}{l|cccc}
\toprule
Method & BLEU@4 & BLEU@1 & ROUGE & CIDEr \\
\midrule
w/o SSL & 17.6 & 6.5 & 30.9 & 24.0 \\
w/ SSL & 18.2 & 7.8 & 32.5 & 24.2 \\
\bottomrule
\end{tabular}
\label{suppl-text}
\end{table}

\paragraph{Effect of SSL on Future Motion Understanding}
We conduct an ablation study to examine how our self-supervised training stage influences the model’s ability to understand future motion semantics. As shown in \cref{suppl-text}, removing the SSL stage and directly training motion prediction and semantic understanding in a joint supervised manner (w/o SSL) leads to worse text generation performance. These results demonstrate that the SSL stage effectively enhances semantic reasoning about future dynamics and provides more informative motion representations for downstream captioning.

\begin{table}[t]
\centering
\setlength\tabcolsep{4.8pt}
% \scriptsize
\small
\caption{Ablation study of model architecture}
\begin{tabular}{ c | c | c c c c} 
\toprule

ID &  & 80ms & 160ms & 320ms & 400ms \\ 
\hline
A & add & 9.4 & 20.1 & 42.7 & 53.5 \\
B & connect & 9.2 & 20.1 & 42.7 & 53.7 \\ 
\hline
C & parallel \textit{attn.} & 9.1 & 20.1 & 42.3 & 52.9 \\
\hline
D & complete & \textbf{8.8} & \textbf{19.3} & \textbf{41.1} & \textbf{51.8} \\

\bottomrule
\end{tabular}
\label{suppl-fusion}
\end{table}

\paragraph{Way of Fusing Past Motion and Future Motion}
In the default setting, we use cross-attention to integrate features of past and future motions. We experimented with other fusion methods, with results shown in \cref{suppl-fusion}: A) Direct addition of features from past and future actions; B) Concatenation of both in the feature dimension, followed by a Linear layer to reduce the number of channels back to the original value. We observed that cross-attention is the best choice because it explicitly computes the temporal and spatial interactions between the past and future motion.

\begin{table}[t]
\centering
\setlength\tabcolsep{3.8pt}
% \scriptsize
\small
\caption{Impact of model size. \(d\) is the feature dimension. \(l\) is the number of layers.}
\begin{tabular}{ c | c | c c c c | c} 
\toprule

ID &  & 80ms & 160ms & 320ms & 400ms & Param\\ 
\hline
E & \( d = 64 \) & 9.3 & 20.2 & 43.3 & 54.4 & 0.67M \\
F & \( d = 128 \) & 9.1 & 19.8 & 42.0 & 52.9 & 1.00M \\
G & \( l = 2 \) & 9.2 & 20.2 & 42.6 & 53.3 & 1.16M \\
H & \( l = 4 \) & \textbf{8.8} & 19.3 & 41.6 & 52.3 & 2.15M \\
\hline
D & complete & \textbf{8.8} & \textbf{19.3} & \textbf{41.1} & \textbf{51.8} & 1.66M \\

\bottomrule
\end{tabular}
\label{suppl-model-size}
\end{table}

\paragraph{Spatiotemporal Attention}
In our default configuration, we adopt the sequential computation scheme from AuxFormer~\cite{xu2023auxiliary}, where temporal attention and spatial attention are applied one after the other to progressively capture motion dynamics and joint dependencies. In contrast, the parallel computation variant (see \cref{suppl-fusion} C), which processes temporal and spatial relations simultaneously, produces noticeably inferior performance, indicating that the sequential formulation provides a more effective inductive bias for structured human motion modeling.

\paragraph{Model Size}
In \cref{suppl-model-size}, we show the impact of different model sizes (including feature dimension \( d \) in attention layer and number of layers \( l \)) on parameter count and results. Although reducing the model size can decrease the number of parameters, it can also lead to decreased performance. We observe that the default setting (\( d \)=256, \( l \)=3) represents the optimal balance between parameter count and performance.

\begin{figure*}[t!]
\centering
\includegraphics[width=0.97\textwidth]{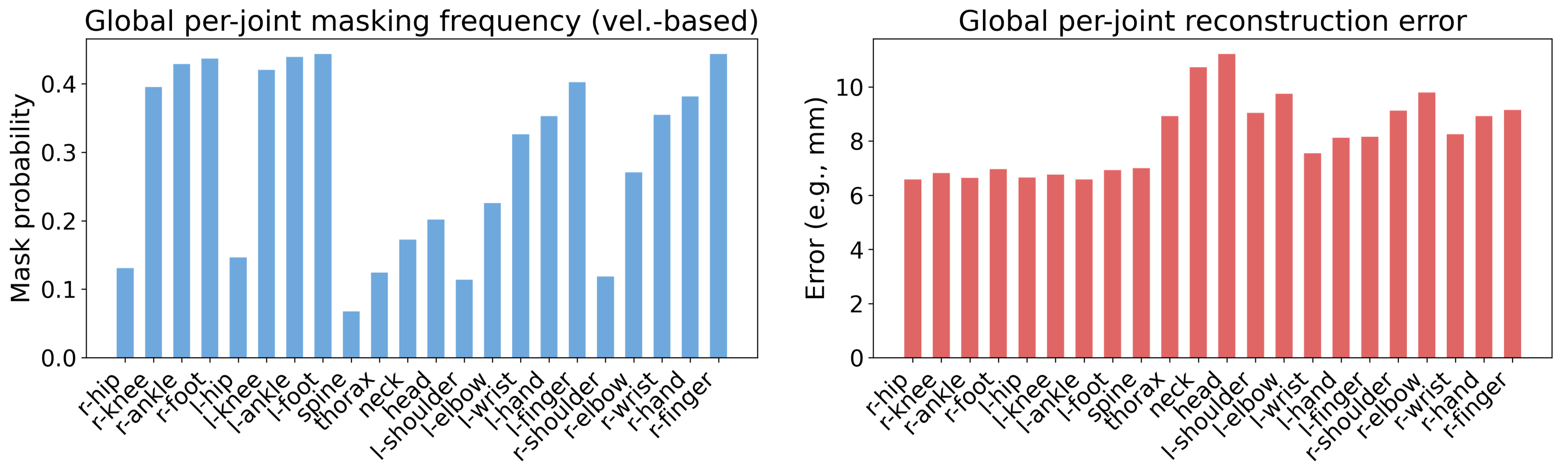}
\caption{Visualization of global per-joint masking probability and the corresponding reconstruction error.}
\label{suppl-mask}
\end{figure*}

\begin{figure*}
\centering
\includegraphics[width=0.97\textwidth]{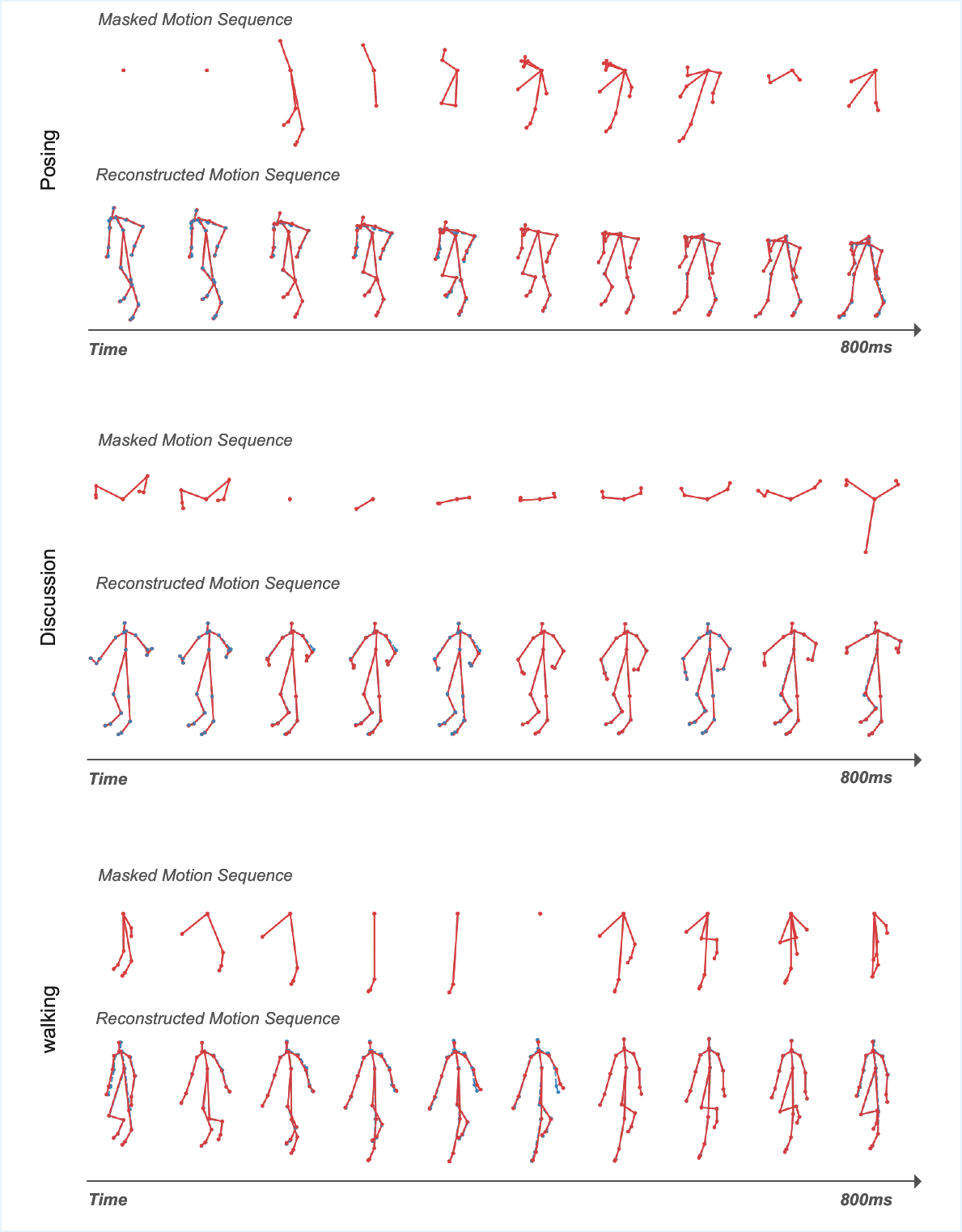}
\caption{Visualization of reconstruction results. The blue lines are ground-truth value and the red lines represent the reconstruction motion skeleton sequences.}
\label{suppl-vis-recon}
\end{figure*}

\begin{table}[t]
\centering
\setlength\tabcolsep{5pt}
\small
\caption{Per-action reconstruction MPJPE at different time horizons (mm).}
\begin{tabular}{lcccc}
\toprule
Action & 80ms & 160ms & 320ms & 400ms \\
\midrule
walking        & 2.55 &  6.25 &  12.54 & 15.48 \\
eating         & 1.51 &  3.47 &  7.04  & 8.81  \\
smoking        & 1.57 &  3.42 &  6.17  & 7.65  \\
discussion     & 1.90 &  4.33 &  7.66  & 9.54  \\
directions     & 1.45 &  3.41 &  7.39  & 9.45  \\
greeting       & 2.45 &  6.08 &  12.52 & 15.64 \\
phoning        & 1.88 &  4.47 &  9.28  & 11.63 \\
posing         & 2.18 &  4.99 &  8.84  & 11.12 \\
purchases      & 2.31 &  5.44 &  11.37 & 14.80 \\
sitting        & 2.02 &  4.65 &  9.63  & 12.30 \\
sittingdown    & 2.50 &  4.86 &  7.91  & 9.83  \\
takingphoto    & 1.81 &  4.17 &  8.93  & 11.59 \\
waiting        & 1.92 &  4.52 &  9.19  & 11.7  \\
walkingdog     & 4.16 & 10.37 &  19.73 & 24.23 \\
walkingtogether& 1.91 &  4.89 &  10.92 & 14.00 \\
\midrule
Average        & 2.14 &  5.02 &  9.94  & 12.82 \\
\bottomrule
\end{tabular}
\label{suppl-recon-mpjpe-per-action}
\end{table}

\section{Analysis of Velocity-Based Joint Masking and Reconstruction Performance}
\label{sec:suppl-mask}
To better understand how velocity-driven masking influences representation learning, we compute the global per-joint masking probability and the corresponding reconstruction error across the entire Human3.6M test set. The results are shown in \cref{suppl-mask}. First, the masking distribution (left) exhibits a clear semantic pattern: High-velocity distal joints such as the feet, wrists, hands, and fingers are assigned higher masking probabilities; low-motion central joints, such as the spine, thorax, and neck, receive much smaller masking rates. This confirms that our velocity-based strategy automatically focuses learning pressure on dynamically informative regions of the body, rather than applying a uniform mask. Such non-uniform perturbation encourages the model to learn motion-consistent representations that reflect true physical dynamics, instead of memorizing static structure.

Second, the reconstruction error plot (right) shows a correlation with the masking distribution. Highly dynamic joints not only move faster but also demonstrate larger pose variability, making them inherently more challenging to reconstruct. Interestingly, although these dynamic joints are frequently masked, their reconstruction error does not explode. As shown in the right figure, the model maintains a stable reconstruction range (approximate 6 mm to 12 mm). This is largely because their adjacent, low-velocity joints remain visible and provide strong spatial constraints, allowing the model to leverage local kinematic continuity for recovery. In other words, the structured skeletal graph prevents error propagation and enables the model to infer masked high-frequency motions using nearby static or slow-moving joints.

The qualitative reconstructions in \cref{suppl-vis-recon} further validate the effectiveness of our masking strategy. Even when large portions of the distal limbs are removed, the model is able to faithfully recover the full-body motion trajectory, producing temporally coherent poses that closely match the unmasked ground truth. Notably, complex motion cues—such as arm swings in Posing, subtle conversational gestures in Discussion, and periodic gait cycles in Walking—are accurately restored, demonstrating the model’s ability to infer fine-grained dynamics from limited visible context. These visual results highlight the robustness of our learned representations and showcase how motion structure priors guide reliable completion of challenging high-frequency movements.

The per-action results in \cref{suppl-recon-mpjpe-per-action} show that dynamic actions with large limb motions incur higher reconstruction errors, while static actions remain easy to recover. Despite this variation, our velocity-aware masking maintains stable performance across both low- and high-motion activities.

\begin{figure*}[t!]
\centering
\includegraphics[width=0.97\textwidth]{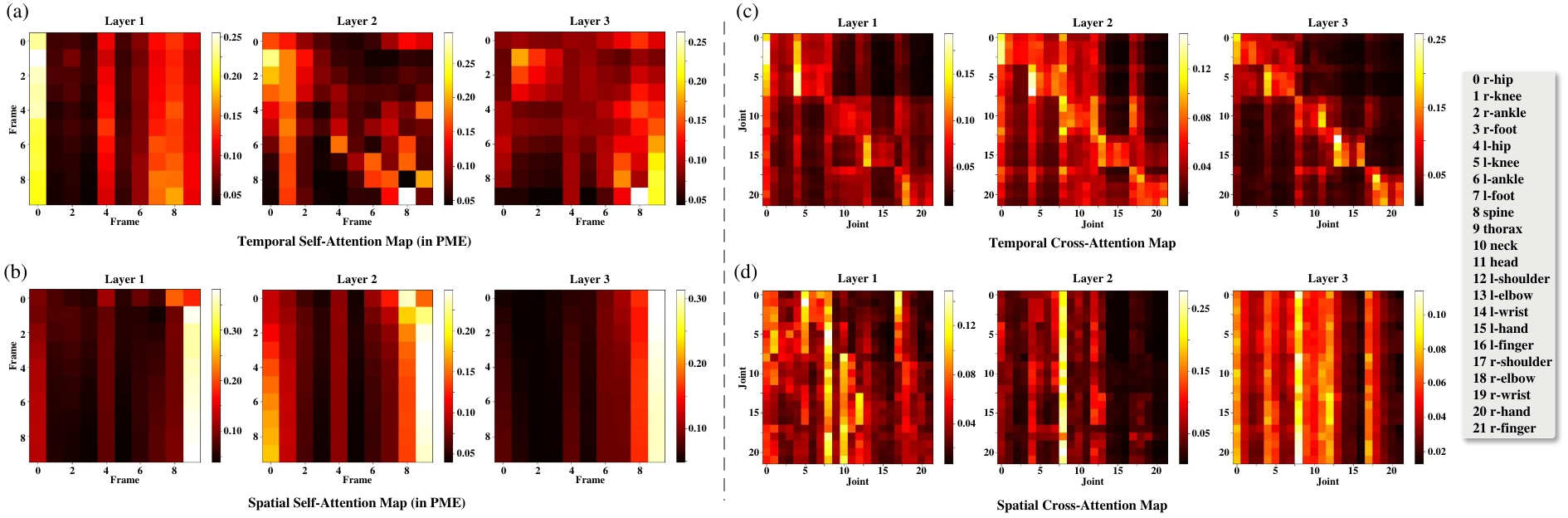}
\caption{Visualization of attention feature maps.}
\label{suppl-attn-map}
\end{figure*}

\section{Visualization of Attention Feature Map}
\label{sec:suppl-vis-attn}
For the example of the Discussion action, we present the averaged attention feature maps over 512 samples. As shown in \cref{suppl-attn-map}, sub figure (a) and (b) display the temporal self-attention and spatial self-attention feature maps from different layers in PME, sub figure (c) and (d) show the temporal cross-attention and spatial cross-attention feature maps from different layers in FMP. 

In temporal cross-attention, the model's first and third layers tend to allocate attention to the later frames, as these frames have a greater influence on future sequences. In the second layer, the model distributes attention to both later and earlier frames, highlighting the potential impact of historical sequences on future actions. In spatial cross-attention, the model identifies significant correlations between the movement patterns of the 'thorax', 'head', 'left elbow', and 'right elbow' joints and the prediction of future actions.

\section{Full Results on Human3.6M}
\label{sec:suppl-result-h36m}
We provide the full results of comparisons on the Human3.6M dataset in \cref{h36m-short-supp} as a supplement to \cref{h36m-short-all}. Despite the simplicity of our network structure, we achieved the best results on most actions.

% \section{Rationale}
\label{supp}

\begin{table*}[t]
\centering
\setlength\tabcolsep{2.5pt}
% \scriptsize
\small
\caption{Comparisons of MPJPEs between our proposed method with other \textit{state-of-the-art} methods for short-term prediction on Human3.6. We show the results across all actions such as walking and eating. The best results are highlighted in bold, and the second best results is underlined}
\begin{tabular}{ c | c c c c | c c c c | c c c c | c c c c | c c c c } 
\toprule

Motion & \multicolumn{4}{c|}{Walking} & \multicolumn{4}{c|}{Eating} & \multicolumn{4}{c|}{Smoking} & \multicolumn{4}{c|}{Discussion} & \multicolumn{4}{c}{Direction} \\
\hline
Milliseconds & 80 & 160 & 320 & 400 & 80 & 160 & 320 & 400 & 80 & 160 & 320 & 400 & 80 & 160 & 320 & 400 & 80 & 160 & 320 & 400 \\ 
\hline
Mixer\cite{bouazizi2022motionmixer} & 12.0 & 23.0 & 42.4 & 51.3 & 7.8 & 16.5 & 34.3 & 43.0 & 7.4 & 15.2 & 30.3 & 37.6 & 10.7 & 25.0 & 56.9 & 71.1 & 7.5 & 18.4 & 43.9 & 55.3 \\
PGBIG\cite{ma2022progressively} & 10.2 & 19.8 & 34.5 & 40.3 & 7.0 & 15.1 & 30.6 & 38.1 & 6.6 & 14.1 & 28.2 & 34.7 & 10.0 & 23.8 & 53.6 & 66.7 & 7.2 & 17.6 & 40.9 & 51.5 \\
SPGSN\cite{li2022skeleton} & 10.1 & 19.4 & 34.8 & 41.5 & 7.1 & 14.8 & 30.5 & 37.9 & 6.7 & 13.8 & 28.0 & 34.6 & 10.4 & 23.8 & 53.6 & 67.1 & 7.4 & 17.2 & \underline{40.0} & \textbf{50.3} \\
siMLPe\cite{guo2023back} & 11.9 & 19.3 & 31.9 & \underline{36.9} & 10.2 & 16.5 & 30.7 & 37.3 & 10.4 & 16.4 & 29.7 & 35.9 & 12.7 & 25.3 & 55.1 & 68.1 & 11.0 & 19.4 & 41.6 & 52.1 \\
DSTD-GC\cite{fu2023learning} & 11.1 & 22.4 & 38.8 & 45.2 & 7.0 & 15.5 & 31.7 & 39.2 & 6.6 & 14.8 & 29.8 & 36.7 & 10.0 & 24.4 & 54.5 & 67.4 & 6.9 & 17.4 & 40.1 & 51.7 \\
AuxFormer\cite{xu2023auxiliary} & 8.9 & \underline{16.9} & \textbf{30.1} & \textbf{36.1} & 6.4 & \underline{14.0} & \textbf{28.8} & \underline{35.9} & 5.7 & \underline{11.4} & \underline{22.1} & \underline{27.9} & \underline{8.6} & \underline{18.8} & \underline{38.8} & \underline{49.2} & 6.8 & 17.0 & 40.3 & 51.6 \\
CIST-GCN\cite{medina2024context} & 11.8 & 23.4 & 40.5 & 46.5 & 6.7 & 14.8 & 29.8 & 36.8 & 7.3 & 15.6 & 31.0 & 38.0 & 10.2 & 23.7 & 52.3 & 65.3 & 7.3 & 18.1 & 43.6 & 55.3 \\
GCNext\cite{wang2024gcnext} & \underline{8.8} & -- & -- & 38.9 & \textbf{5.9} & -- & -- & \textbf{35.0} & \underline{5.6} & -- & -- & 36.1 & 8.8 & -- & -- & 63.1 & -- & -- & -- & -- \\
ALIEN\cite{wei2025alien} & 9.5 & 18.7 & 33.8 & 40.2 & 6.6 & 14.6 & 30.5 & 38.1 & 6.4 & 13.9 & 28.8 & 35.7 & 9.6 & 23.7 & 53.8 & 67.2 & \underline{6.6} & \underline{16.8} & 40.5 & \underline{51.4} \\
Ours & \textbf{8.6} & \textbf{16.5} & \underline{31.2} & 37.7 & \underline{6.1} & \textbf{13.7} & \underline{29.3} & 36.8 & \textbf{5.3} & \textbf{10.3} & \textbf{19.7} & \textbf{25.2} & \textbf{7.7} & \textbf{16.3} & \textbf{31.7} & \textbf{40.9} & \textbf{6.3} & \textbf{15.8} & \textbf{39.8} & \underline{51.4} \\
\hline

\hline
Motion & \multicolumn{4}{c|}{Greeting} & \multicolumn{4}{c|}{Phoning} & \multicolumn{4}{c|}{Posing} & \multicolumn{4}{c|}{Purchases} & \multicolumn{4}{c}{Sitting} \\
\hline
Milliseconds & 80 & 160 & 320 & 400 & 80 & 160 & 320 & 400 & 80 & 160 & 320 & 400 & 80 & 160 & 320 & 400 & 80 & 160 & 320 & 400 \\ 
\hline
Mixer\cite{bouazizi2022motionmixer} & 16.4 & 37.1 & 80.2 & 98.2 & 9.1 & 19.5 & 42.4 & 53.6 & 11.5 & 27.0 & 64.9 & 83.5 & 13.4 & 30.7 & 65.6 & 80.0 & 9.8 & 20.9 & 45.6 & 57.3 \\
PGBIG\cite{ma2022progressively} & 15.2 & 34.1 & 71.6 & 87.1 & 8.3 & 18.3 & 38.7 & 48.4 & 10.7 & 25.7 & 60.0 & 76.6 & 12.5 & 28.7 & \underline{60.1} & \underline{73.3} & 8.8 & \underline{19.2} & 42.4 & 53.8 \\
SPGSN\cite{li2022skeleton} & 14.6 & 32.6 & 70.6 & 86.4 & 8.7 & 18.3 & 38.7 & 48.5 & 10.7 & 25.3 & 59.9 & 76.5 & 12.7 & 28.6 & 61.0 & 74.4 & 9.3 & 19.4 & \underline{42.2} & \underline{53.6} \\
siMLPe\cite{guo2023back} & 16.2 & 33.7 & 70.9 & 86.3 & 11.9 & 20.3 & 39.7 & 48.9 & 13.9 & 27.3 & 62.0 & 78.6 & 15.5 & 29.3 & \textbf{59.3} & \textbf{72.0} & 13.2 & 21.9 & 44.5 & 55.7 \\ 
DSTD-GC\cite{fu2023learning} & 14.3 & 33.5 & 72.2 & 87.3 & 8.5 & 19.2 & 40.3 & 49.9 & 10.1 & 25.4 & 60.6 & 77.3 & 12.7 & 29.7 & 62.3 & 75.8 & 8.8 & 19.3 & 42.9 & 54.3 \\
AuxFormer\cite{xu2023auxiliary} & \underline{13.5} & 31.3 & 69.2 & 85.4 & \underline{7.9} & \underline{17.3} & \textbf{37.4} & \textbf{47.2} & \underline{8.8} & \underline{19.1} & \underline{39.2} & \underline{51.0} & \underline{11.9} & \underline{28.0} & 61.8 & 76.3 & \underline{8.7} & \textbf{19.0} & \textbf{42.1} & \textbf{53.3} \\
CIST-GCN\cite{medina2024context} & 13.7 & \underline{31.0} & \underline{65.7} & \underline{79.9} & 8.6 & 18.5 & 39.3 & 49.6 & 9.6 & 23.7 & 57.7 & 75.0 & 13.3 & 30.2 & 63.0 & 77.3 & 8.9 & 19.4 & 42.3 & \underline{53.6} \\
ALIEN\cite{wei2025alien} & 13.6 & 32.1 & 70.3 & 85.8 & 8.1 & 18.2 & 39.3 & 49.4 & 9.8 & 25.1 & 61.2 & 78.9 & 12.2 & 28.9 & 62.0 & 76.6 & \underline{8.7} & \underline{19.2} & 42.9 & 54.9 \\
Ours & \textbf{10.8} & \textbf{26.4} & \textbf{62.7} & \textbf{79.8} & \textbf{7.5} & \textbf{17.1} & \underline{37.9} & \underline{48.0} & \textbf{8.2} & \textbf{16.9} & \textbf{32.8} & \textbf{43.5} & \textbf{11.4} & \textbf{27.6} & 62.6 & 77.6 & \textbf{8.5} & \underline{19.2} & 44.3 & 56.7 \\
\hline

\hline
Motion & \multicolumn{4}{c|}{Sitting Down} & \multicolumn{4}{c|}{Taking Photo} & \multicolumn{4}{c|}{Waiting} & \multicolumn{4}{c|}{Walking Dog} & \multicolumn{4}{c}{Walking Together} \\
\hline
Milliseconds & 80 & 160 & 320 & 400 & 80 & 160 & 320 & 400 & 80 & 160 & 320 & 400 & 80 & 160 & 320 & 400 & 80 & 160 & 320 & 400 \\
\hline
Mixer\cite{bouazizi2022motionmixer} & 15.3 & 30.5 & 62.7 & 78.0 & 9.0 & 19.9 & 44.7 & 56.9 & 9.9 & 22.0 & 49.7 & 62.5 & 21.1 & 43.9 & 84.7 & 101.2 & 9.9 & 20.4 & 39.3 & 47.3 \\
PGBIG\cite{ma2022progressively} & 13.9 & 27.9 & 57.4 & 71.5 & 8.4 & 18.9 & 42.0 & 53.3 & 8.9 & 20.1 & 43.6 & 54.3 & \underline{18.8} & \underline{39.3} & \underline{73.7} & 86.4 & 8.7 & 18.6 & 34.4 & 41.0 \\
SPGSN\cite{li2022skeleton} & 14.2 & 27.7 & \underline{56.7} & 70.7 & 8.8 & 18.9 & \underline{41.5} & \underline{52.7} & 9.2 & 19.8 & 43.1 & 54.1 & -- & -- & -- & -- & 8.9 & 18.2 & 33.8 & 40.9 \\
siMLPe\cite{guo2023back} & 18.0 & 29.8 & 57.7 & 71.2 & 13.1 & 21.3 & 42.9 & 53.6 & 11.9 & 21.5 & 44.4 & 55.0 & 21.8 & 40.6 & 74.6 & 87.0 & 11.9 & 19.8 & 34.2 & 39.8 \\
DSTD-GC\cite{fu2023learning} & 14.1 & 28.0 & 57.3 & 71.2 & 8.4 & 18.8 & 42.0 & 53.5 & 8.7 & 20.2 & 44.3 & 55.3 & 19.6 & 41.8 & 77.6 & 90.2 & 9.1 & 19.8 & 36.3 & 42.7 \\
AuxFormer\cite{xu2023auxiliary} & 13.5 & 27.6 & 57.7 & 72.2 & \underline{8.2} & \underline{18.4} & \underline{41.5} & 53.0 & \underline{8.2} & \underline{18.5} & \textbf{41.2} & \textbf{52.2} & \textbf{17.1} & \textbf{36.5} & \textbf{70.4} & \textbf{83.0} & \underline{7.8} & \underline{15.9} & \textbf{30.2} & \textbf{37.0} \\
CIST-GCN\cite{medina2024context} & 14.1 & 29.8 & 57.3 & \underline{69.8} & \underline{8.2} & \underline{18.4} & \textbf{40.6} & \textbf{51.8} & 8.6 & 19.4 & 43.5 & 54.8 & 20.0 & 41.4 & \underline{73.7} & \underline{85.1} & 9.6 & 20.3 & 38.2 & 45.6 \\
ALIEN\cite{wei2025alien} & \underline{13.4} & \underline{27.0} & 56.4 & 70.5 & 8.8 & 19.0 & 42.1 & 53.5 & 8.3 & 19.2 & 43.3 & 54.8 & 18.7 & 40.3 & 75.9 & 88.5 & 8.4 & 17.8 & 33.7 & 40.4 \\
Ours & \textbf{10.1} & \textbf{17.2} & \textbf{30.2} & \textbf{39.2} & \textbf{7.6} & \textbf{17.8} & 42.0 & 54.1 & \textbf{7.8} & \textbf{18.0} & \underline{41.6} & \underline{53.2} & 19.1 & 41.1 & 80.6 & 95.2 & \textbf{7.3} & \textbf{15.3} & \underline{30.8} & \underline{37.9} \\

\bottomrule
\end{tabular}
\label{h36m-short-supp}
\end{table*}

\end{document}